\documentclass{article}
\usepackage{geometry}
\usepackage{tikz}
\usepackage{cite}
\usetikzlibrary{calc}
\usetikzlibrary{arrows,shapes,chains}
\usepackage{xcolor}
\usepackage{amsfonts}
\usepackage{amsmath}
\usepackage{graphicx}
\usepackage{multirow}
\usepackage{algorithmic}
\usepackage{algorithm}
\usepackage{subfigure}
\usepackage{epstopdf}
\usepackage{url}
\usepackage{cleveref}
\crefname{equation}{}{}
\crefname{table}{TABLE}{TABLES}
\crefname{figure}{Fig.}{Figs.}
\crefname{section}{Section}{Sections}

\DeclareMathOperator*{\argmin}{arg\,min}

\newcommand{\cb}[1]{\ifmmode {\boldsymbol{#1}}\else ${\boldsymbol{#1}}$\fi}
\newcommand{\tabincell}[2]{\begin{tabular}{@{}#1@{}}#2 \end{tabular}}
\newcommand{\cp}[1]{\ifmmode {\mathcal{#1}}\else ${\mathcal{#1}}$\fi}

\title{A CNN-based Spatial Feature Fusion Algorithm for Hyperspectral Imagery Classification}
\author{Alan J.X. Guo~\textit{Member,~IEEE} and Fei~Zhu~\textit{Member,~IEEE}
	\thanks{A.~Guo and F.~Zhu are with the Center for Applied Mathematics, Tianjin University, China. (jiaxiang.guo;~fei.zhu@tju.edu.cn) }
}

\begin{document}
\maketitle
\begin{abstract}
The shortage of training samples remains one of the main obstacles in applying the neural networks to the hyperspectral images classification. To fuse the spatial and spectral information, pixel patches are often
utilized to train a model, which may further aggregate this problem.
In the existing works, an ANN model supervised
by center-loss (ANNC) was introduced. Training merely with spectral information, the ANNC yields
discriminative spectral features suitable for the subsequent classification tasks.
In this paper, we propose a novel CNN-based spatial feature fusion (CSFF) algorithm which allows
a smart integration of spatial information to the spectral features extracted by ANNC. As a critical part of CSFF,
a CNN-based discriminant model is introduced to estimate whether two pixels belong to the same class.
At the testing stage, by applying the discriminant model to the pixel pairs generated by a test pixel and each of its neighbors, the local structure
is estimated and represented as a customized convolutional kernel. The spectral-spatial feature is generated by
a convolutional operation between the estimated kernel and the corresponding spectral features within a local region. The final label is determined by classifying the resulting spectral-spatial feature.
Without increasing the number of training
samples or involving pixel patches at the training stage, the CSFF framework achieves the state-of-the-art
by declining $20\%-50\%$ classification failures in experiments
on three well-known hyperspectral images.
\end{abstract}
	
\section{Introduction}
A hyperspectral image is a collection of spectral pixels. Each of them records a
continuous reflection
spectrum over a same land-cover, with hundreds of channels across a certain wavelength range.
The classification of these spectral pixels into a set of land-cover materials is a crucial task in hyperspectral images analysis\cite{lu2007survey}.
To address this issue, an amount of spectrum-based works are proposed mainly by exploiting the abundant spectral information containing in the data.
Among, linear algorithms in dimension reduction and feature classification are
the most investigated, for example, principal component analysis (PCA) \cite{licciardi2012linear}, independent component analysis (ICA) \cite{villa2011hyperspectral} and linear discriminant analysis (LDA) \cite{bandos2009classification}.
Nonlinear models and their extensions are also introduced to achieve better representations of the spectra, including but not
limited to support vector machine (SVM) \cite{melgani2004classification}, manifold learning \cite{lunga2014manifold}, random forest \cite{ham2005investigation}
and kernel-based strategies \cite{kuo2009kernel,scholkopf2002learning}.

Benefiting the increasing imaging qualities of both spectral and spatial resolutions \cite{fauvel2013advances}, numerous
spectral-spatial algorithms have been developed in order to obtain more accurate classification performance \cite{chen2011hyperspectral,fauvel2008spectral,huo2011spectral,li2013spectral,plaza2009incorporation,song2014remotely,Hang2017Robust}.
Specifically, a multi-scale adaptive sparse representation (MASR) method is proposed in \cite{Fang2014Spectral},
where the spatial information at different scales is explored simultaneously.
In MASR,
each neighboring pixel is represented via an adaptive sparse combination of training samples,
and the label of the centering test pixel is determined by the recovered sparse coefficients.
It is noteworthy that although MASR utilizes the spatial information at the testing stage,
only spectral information is engaged at the training stage.
This helps to mitigate the shortage of training data to some degree.

Deep learning frameworks, including artificial neural networks (ANN), convolutional neural networks (CNN), and recurrent neural network (RNN) have been successfully applied in many fields related to machine learning and signal processing \cite{Hill1994Artificial,zhang2000neural,krizhevsky2012imagenet,Schmidhuber2015deep,hochreiter1997long}.
Recently, the
neural network-based
models have been utilized in hyperspectral images classification,
achieving remarkable improvements over the
traditional methods in terms of classification performance.
Earlier works include the stacked autoencoder (SAE)~\cite{benediktsson2005classification,chen2014deep}, the deep belief network (DBN)~\cite{chen2015spectral} and {\em etc.}.
More recently, many researches are dedicated to the varieties of CNN and RNN-based models, as studied in~\cite{slavkovikj2015hyperspectral,chen2016deep,liang2016hyperspectral,  zhao2016spectral,romero2016unsupervised,yu2017convolutional,Jiao2017Deep,Nogueira2017Towards,guo2017spectral,Mou2017Deep, Liu2017Bidirectional}.

Training with pixel patches is a natural idea to take advantage of
both spectral and spatial information, and is adopted by most of the aforementioned
neural network-based studies
\cite{slavkovikj2015hyperspectral,chen2016deep,liang2016hyperspectral,  zhao2016spectral,romero2016unsupervised,yu2017convolutional,Jiao2017Deep,Nogueira2017Towards,guo2017spectral,Liu2017Bidirectional}.
Representative works include the so-called $3$D-CNN in~\cite{chen2016deep}, where pixel patches are directly fed to the deep model, and the integrated spectral-spatial features can be extracted from the hyperspectral data. 
However, it should be noticed that this strategy may further aggravate the shortage of training data.
Different from the numerous accessible RGB images, the labeled samples are
limited in hyperspectral imagery, which are usually insufficient for network training \cite{chen2016deep}.
Compared with the spectrum-based models, the pixel-patch-based ones aggregate this contradiction from two aspects,
\begin{itemize}
	\item using pixel patches often complicates the model by introducing additional undetermined parameters,
	thus requiring more training data;
	\item a hyperspectral image usually contains fewer mutually non-overlapped patches with specified size than
	pixel-wise samples.
\end{itemize}
Several attempts have been made to overcome the shortage of training data,
such as the virtual sample strategy \cite{chen2016deep},
the pre-training and fine-tuning techniques \cite{Jiao2017Deep,Nogueira2017Towards}.

Moreover, to circumvent the obstacle raised by training with pixel patches, frameworks are built by concatenating
a spectrum-based model and a post spatial information integration scheme.
Following this idea, the pixel-pair feature algorithm (PPF) \cite{li2017hyperspectral} and the
ANNC with adaptive spatial-spectral center classifier
(ANNC-ASSCC) \cite{guo2017spectral} have been proposed in
the literature.
Together with the aforementioned MASR and $3$D-CNN, the four methods are chosen to be compared
with the proposed CSFF algorithm in the experiments,
as to be reported in \cref{sec: Experiments}.

The PPF framework investigates
a pixel-pair-based classification model based on CNN.
Let $x_1$ and $x_2$ be two pixels with label $\ell(x_1)$ and $\ell(x_2)$, respectively. The pixel-pair for training is generated as $(x_1,x_2)$, with the label determined by
\begin{equation*}
	\ell((x_1,x_2))=
		\begin{cases}
		\ell(x_1) & \textrm{ if } \ell(x_1) = \ell(x_2),\\
		0& \textrm{ otherwise. }
		\end{cases}
\end{equation*}
It is observed that the number of pixel-pairs computes
$\binom{N}{2}$, where $N$ is the size of the training set.
This enlarged size of the training set enables the training of a deep CNN model.
The model classifies the pixel-pairs into $K+1$ classes with labels $0$ and $1,2,\ldots,K$,
where $K$ is the number of pixel classes.
At the testing stage, pixel-pairs are firstly generated by the test pixel and its neighbors,
and then classified into $K+1$ classes by the learned CNN model.
The final label of a test pixel is determined by a voting strategy among the non-zero classification results.

\begin{figure}
{\linespread{1}
	\centering
	\scriptsize
	\tikzstyle{format}=[circle,draw,thin,fill=white]
	\tikzstyle{format_gray}=[circle,draw,thin,fill=gray]
	\tikzstyle{format_rect}=[rectangle,draw,thin,fill=white,align=center]
	\tikzstyle{arrowstyle} = [->,thick]
	\scalebox{0.7}
	{
		\begin{tikzpicture}[node distance=4mm,  auto,>=latex',  thin,  start chain=going below, every join/.style={norm}, scale=0.3]
		\definecolor{gray_so}{RGB}{88,110,117}
		\definecolor{yellow_so}{RGB}{181,137,0}
		\definecolor{cyan_so}{RGB}{42,161,152}
		\definecolor{orange_so}{RGB}{203,75,22}
		\filldraw[fill=yellow_so,rounded corners,fill opacity=0.33,style=dashed] (-2,-2) rectangle (2,15);
		\foreach \x/\xtext in {0, 2, 4, 6, 8, 10, 12, 14, 16, 18}
		{
			\node at (0,\x/4+3.75) [format] { };
			\node at (0,\x/4+0.25+3.75) [format_gray] { };
		}
		\filldraw[fill=orange_so,rounded corners,fill opacity=0.2,style=dashed] (3,-2) rectangle (17,15);
		\foreach \x/\xtext in {0,2,4,6,8,10,12,14,16,18,20,22,24,26,28,30,32,34,36,38,40,42,44,46,48}
		{
			\node at (5,\x/4) [format] { };
			\node at (5,\x/4+0.25) [format_gray] { };
		}
		\filldraw[fill=cyan_so,rounded corners,fill opacity=0.33,style=dashed] (18,-2) rectangle (27,15);
		\foreach \x/\xtext in {0,2,4,6,8,10,12,14,16,18,20,22,24,26,28}
		{
			\node at (10,\x/4+2.5) [format] { };
			\node at (10,\x/4+0.25+2.5) [format_gray] { };
		}
		\foreach \x/\xtext in {0, 2, 4, 6, 8, 10, 12}
		{
			\node at (15,\x/4+4.5) [format] { };
			\node at (15,\x/4+0.25+4.5) [format_gray] { };
		}
		\foreach \x/\xtext in {0, 2, 4, 6}
		{
			\node at (20,\x/4+5) [format_gray] { };
			\node at (20,\x/4+0.25+5) [format] { };
		}
		\node at (20,8/4+5) [format_gray] { };
		
		\foreach \x/\xtext in {0, 2, 4, 6}
		{
			\node at (25,\x/4+5) [format_gray] { };
			\node at (25,\x/4+0.25+5) [format] { };
		}
		\node at (25,8/4+5) [format_gray] { };
		
		\draw[arrowstyle] (1,6.25) -- node {FC} (4,6.25);
		\draw[arrowstyle] (6,6.25) --  node {FC} (9,6.25);
		\draw[arrowstyle] (11,6.25) -- node {FC} (14,6.25);
		\draw[arrowstyle] (16,6.25) -- node {FC} (19,6.25);
		\draw[arrowstyle] (21,6.25) -- node {Softmax} (24,6.25);
		\node at (0,0) (n0) {};
		\node at (5,0) (n5) {};
		\node at (10,0) (n10) {};
		\node at (15,0) (n15) {};
		\node at (20,0) (n20) {};
		\node at (25,0) (n25) {};
		\node[below of =n0] (i) {Input};
		\node[below of =n5] (l1) {Layer 1};
		\node[below of =n10] (l2) {Layer 2};
		\node[below of =n15] (l3) {Layer 3};
		\node[below of =n20] (o) {Output};
		\node[below of =n25] (label) {Label};
		\node at (15,12) [format_rect] (centerloss) {center loss};
		\node at (25,12) [format_rect] (loss) {softmax loss};
		\draw[arrowstyle] (15,8.5) to (centerloss.south);
		\draw[arrowstyle] (25,7.75) to (loss.south);
		\node at (10,14) [rounded corners,style=dashed,color=gray_so] (input_section) {Feature Extraction Network};
		\node at (22.5,14) [rounded corners,style=dashed,color=gray_so] (input_section) {Logistic Regression};
		\end{tikzpicture}
	}
	\caption{\label{mlp_struct} Structure of the spectral feature extracting network~\cite{guo2017spectral}.}
}
\end{figure}
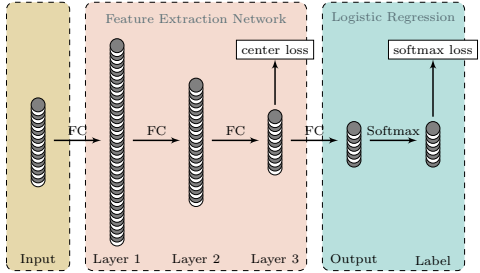

In \cite{guo2017spectral}, an ANN-based feature extraction and classification framework is proposed.
The so-called ANNC-ASSCC algorithm consists of two successive steps:
\begin{itemize}
	\item spectral feature extraction with ANN;
	\item adaptive spatial information fusion and classification.
\end{itemize}
In the first step, a simple but efficient ANN structure is designed to extract spectral features.
As illustrated in \cref{mlp_struct},
the network contains $4$ fully connected layers, and a joint supervision of softmax loss and center loss
is applied for classification task.
During the training stage, the $k$ class centers, {\em i.e.} $c_1, c_2, \ldots c_k$, are updated by averaging
the $3$-rd layer's outputs within respective classes.
For a training pixel, the introduced center loss encourages
the output of the $3$-rd layer to gather around its corresponding class center in Euclidean space.
At the testing stage, the first $3$ layers of the learned network are utilized and the outputs of the $3$-rd layer are regarded as
the extracted spectral features.
In the second step, the spatial
information is merged by averaging the extracted spectral features within
neighboring regions of different sizes.
After the resulting features are classified, a voting strategy is applied to generate the final prediction.

In this paper, we choose the ANNC network in \cite{guo2017spectral} as the spectral feature extraction model, and design
a CNN-based Spacial Feature Fusion (CSFF) algorithm to posteriorly integrate the spatial information.
The main characters of the proposed classification framework are listed as below.
\begin{enumerate}
    \item Rather than implicit spatial information utilized in \cite{guo2017spectral}, explicit spatial structures are produced by our novel CNN-based model.
    Therefore, the CSFF works more reasonably than the averaging fusion strategy in the previous work \cite{guo2017spectral}.
    \item The CSFF algorithm shares the same training data with the spectral feature extraction model, which is chosen as ANNC in this paper. Without increasing the size of the training set, this settlement keeps the
    whole framework working on a small amount of training data.
    \item To enhance the representative ability of the framework, the spatial structure extraction algorithm
    in CSFF is
    designed to have a totally different network structure from ANNC, such that
    the two models are distinguished at feature expression and abstraction levels.
    To be precise, the ANNC expresses the shallow features of spectra, while the
    spatial structure extraction algorithm
    produces more abstract features with a deeper and more complex model structure.
    This helps to alleviate the correlation between these
    two models, thus increasing the performance of the whole framework.
\end{enumerate}
Taking advantage of the above characters, without requiring more training data,
the proposed CSFF algorithm allows to achieve the state-of-the-art by
declining $20\%-50\%$ classification failures in experiments.

The reminder of this paper is organized as follows. The whole framework outline and the CSFF algorithm are
discussed in \cref{sec:ProposedModel}.
\cref{sec: Experiments} reports the experimental results and analysis. In \cref{sec: Conclusion},
some concluding remarks are presented.

\section{Framework outline and the CNN-based spatial feature fusion algorithm}\label{sec:ProposedModel}

In this section, we present an end-to-end spectral-spatial feature extraction and classification framework.
The whole structure is firstly outlined, with detailed explanations on each of the three components.
As a key ingredient of the whole framework, the CSFF algorithm is introduced with emphasis placed on
the newly proposed CNN-based
discriminant model, which outputs the predicted probability of two pixels belonging to the same class.

\subsection{Framework outline}
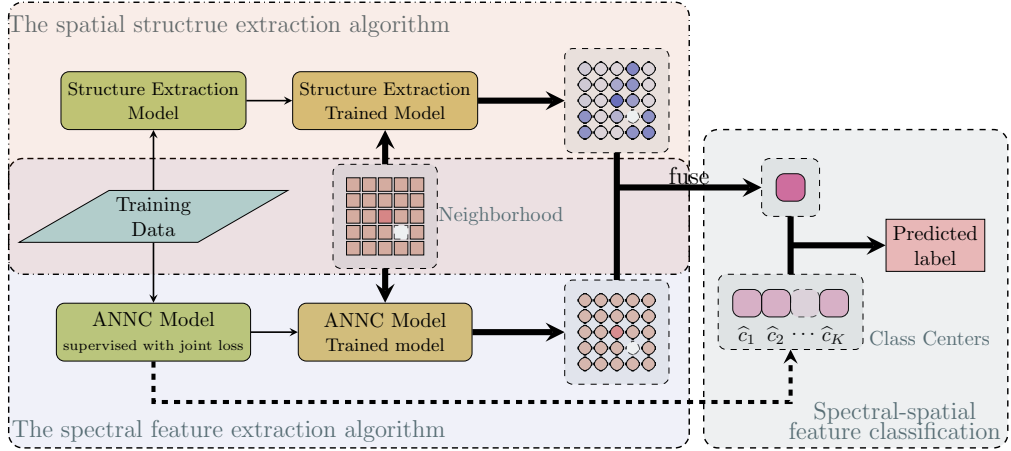
\begin{figure*}
{\linespread{1}
	\centering
	\tikzstyle{format}=[circle,draw,thin,fill=white]
	\tikzstyle{format_gray}=[circle,draw,thin,fill=gray]
	\tikzstyle{format_rect}=[rectangle,draw,thin,fill=white,align=center]
	\tikzstyle{arrowstyle} = [->,thick]
	\tikzstyle{network} = [rectangle, minimum width = 3cm, minimum height = 1cm, text centered, draw = black,align=center,rounded corners,fill=green_so,fill opacity=0.5,text opacity=1]
	\tikzstyle{training_batch} = [trapezium, trapezium left angle = 30, trapezium right angle = 150, minimum width = 3cm, text centered, draw = black, fill = cyan_so, fill opacity=0.3,text opacity=1,align=center]		
	\tikzstyle{class_features} = [trapezium, trapezium left angle = 30, trapezium right angle = 150, minimum width = 3cm, text centered, draw = black, fill = cyan_so, fill opacity=0.3,text opacity=1,align=center]
	\tikzstyle{pixel} = [rectangle, draw = black, fill = orange_so, fill opacity=0.3,text opacity=0,align=center]	
	\tikzstyle{feature} = [rectangle, draw = black, fill = orange_so, fill opacity=0.3,text opacity=0,align=center,rounded corners]	
	\tikzstyle{feature_sfp} = [rectangle, draw = black, fill = violet_so, fill opacity=0.3,text opacity=0,align=center,rounded corners]					
	\tikzstyle{arrow1} = [thick, ->, >= stealth]
	\tikzstyle{arrow1_thick} = [thick, ->, >= stealth, line width=3pt]
	\tikzstyle{arrow2} = [thick, dashed, ->, >= stealth]
	\begin{tikzpicture}[auto,>=latex',  thin,  start chain=going below, every join/.style={norm}]
	\definecolor{gray_so}{RGB}{88,110,117}
	\definecolor{yellow_so}{RGB}{181,137,0}
	\definecolor{cyan_so}{RGB}{42,161,152}
	\definecolor{orange_so}{RGB}{203,75,22}
	\definecolor{green_so}{RGB}{133,153,0}
	\definecolor{red_so}{RGB}{220,50,47}
	\definecolor{magenta_so}{RGB}{211,54,130}
	\definecolor{violet_so}{RGB}{108,113,196}
	\useasboundingbox (0,0) rectangle (18*0.77,8*0.77);
	\scope[transform canvas={scale=0.77}]

	\coordinate (zero_1) at (0,0);

	\node at ($(zero_1)+(4.3,7.3)$) [color=gray_so]{\large{The spatial structrue extraction algorithm}};
	\filldraw[fill=violet_so,rounded corners=7,fill opacity=0.1,style=densely dashed,line width=0.7] ($(zero_1)+(0.5,5)$) rectangle ($(zero_1)+(12.25,0)$);
	\filldraw[fill=orange_so,rounded corners=7,fill opacity=0.1,style= dashdotted,line width=0.7] ($(zero_1)+(0.5,3)$) rectangle ($(zero_1)+(12.25,7.7)$);
	\node at ($(zero_1)+(4.3,0.3)$) [color=gray_so]{\large{The spectral feature extraction algorithm}};
	\filldraw[fill=gray_so,rounded corners=7,fill opacity=0.1,style=dashed,line width=0.7] ($(zero_1)+(12.5,0)$) rectangle ($(zero_1)+(17.75,5.5)$);
	\node at ($(zero_1)+(15.8,0.6)$) [color=gray_so]{\large{Spectral-spatial}};
	\node at ($(zero_1)+(15.8,0.3)$) [color=gray_so]{\large{feature classification}};
	\node(training_data) at ($(zero_1)+(3,4)$)[training_batch] {Training \\Data};
	\node(struct_model_pretrain) at ($(training_data)+(0,2)$)[network] {\small{Structure Extraction}\\ \small{Model}};
	\node(annc_pretrain) at ($(training_data)+(0,-2)$)[network] {ANNC Model \\\scriptsize{supervised with joint loss}};
	\node(struct_model_trained) at ($(struct_model_pretrain)+(4,0)$)[network,fill=yellow_so] {\small{Structure Extraction}\\ \small{Trained Model}};
	\node(annc_trained) at ($(annc_pretrain)+(4,0)$)[network,fill=yellow_so] {ANNC Model \\\small{Trained model}};
	\coordinate(centers) at ($(annc_pretrain)+(0,-1.5)$);

	\draw[arrow1] (training_data.north) -- (struct_model_pretrain.south);
	\draw[arrow1] (training_data.south) -- (annc_pretrain.north);
	
	\draw[arrow1] (struct_model_pretrain.east) -- (struct_model_trained.west);
	\draw[arrow1] (annc_pretrain.east) -- (annc_trained.west);
	\node(p10)[pixel,fill=red_so,fill opacity=0.5] at ($(training_data)+(4,0)$) {};
	\node(p11)[pixel] at ($(p10)-(0.275,0.275)$) {};
	\node(p12)[pixel] at ($(p10)-(0.275,0)$) {};
	\node(p13)[pixel] at ($(p10)-(0.275,-0.275)$) {};
	\node(p14)[pixel] at ($(p10)-(0,0.275)$) {};
	\node(p15)[pixel] at ($(p10)-(0,-0.275)$) {};
	\node(p16)[pixel,fill opacity=1,fill=white,style=dashed,draw opacity=0.7] at ($(p10)-(-0.275,0.275)$) {};
	\node(p17)[pixel] at ($(p10)-(-0.275,0)$) {};
	\node(p18)[pixel] at ($(p10)-(-0.275,-0.275)$) {};
	\node(p19)[pixel] at ($(p10)-(0.55,0.55)$) {};
	\node(p110)[pixel] at ($(p10)-(0.55,0.275)$) {};
	\node(p111)[pixel] at ($(p10)-(0.55,0)$) {};
	\node(p112)[pixel] at ($(p10)-(0.55,-0.275)$) {};
	\node(p113)[pixel] at ($(p10)-(0.55,-0.55)$) {};
	\node(p114)[pixel] at ($(p10)-(0.275,0.55)$) {};
	\node(p115)[pixel] at ($(p10)-(0.275,-0.55)$) {};
	\node(p116)[pixel] at ($(p10)-(0,0.55)$) {};
	\node(p117)[pixel] at ($(p10)-(0,-0.55)$) {};
	\node(p118)[pixel] at ($(p10)-(-0.275,0.55)$) {};
	\node(p119)[pixel] at ($(p10)-(-0.275,-0.55)$) {};
	\node(p120)[pixel] at ($(p10)-(-0.55,0.55)$) {};
	\node(p121)[pixel] at ($(p10)-(-0.55,0.275)$) {};
	\node(p122)[pixel] at ($(p10)-(-0.55,0)$) {};
	\node(p123)[pixel] at ($(p10)-(-0.55,-0.275)$) {};
	\node(p124)[pixel] at ($(p10)-(-0.55,-0.55)$) {};
	\filldraw[fill=gray_so,rounded corners,fill opacity=0.1,style=dashed] ($(p10)+(-0.9,-0.9)$) rectangle ($(p10)+(0.9,0.9)$);
	\node at ($(p10)+(2,0)$) [rounded corners,style=dashed,color=gray_so] {Neighborhood};
	\draw[arrow1_thick] ($(p10)+(0,0.9)$) -- (struct_model_trained.south);
	\draw[arrow1_thick] ($(p10)+(0,-0.9)$) -- (annc_trained.north);
	\node(fp10)[feature,fill=red_so,fill opacity=0.5] at ($(annc_trained)+(4,0)$) {};
	\node(fp11)[feature] at ($(fp10)-(0.275,0.275)$) {};
	\node(fp12)[feature] at ($(fp10)-(0.275,0)$) {};
	\node(fp13)[feature] at ($(fp10)-(0.275,-0.275)$) {};
	\node(fp14)[feature] at ($(fp10)-(0,0.275)$) {};
	\node(fp15)[feature] at ($(fp10)-(0,-0.275)$) {};
	\node(fp16)[feature,fill opacity=1,fill=white,style=dashed,draw opacity=0.7] at ($(fp10)-(-0.275,0.275)$) {};
	\node(fp17)[feature] at ($(fp10)-(-0.275,0)$) {};
	\node(fp18)[feature] at ($(fp10)-(-0.275,-0.275)$) {};
	\node(fp19)[feature] at ($(fp10)-(0.55,0.55)$) {};
	\node(fp110)[feature] at ($(fp10)-(0.55,0.275)$) {};
	\node(fp111)[feature] at ($(fp10)-(0.55,0)$) {};
	\node(fp112)[feature] at ($(fp10)-(0.55,-0.275)$) {};
	\node(fp113)[feature] at ($(fp10)-(0.55,-0.55)$) {};
	\node(fp114)[feature] at ($(fp10)-(0.275,0.55)$) {};
	\node(fp115)[feature] at ($(fp10)-(0.275,-0.55)$) {};
	\node(fp116)[feature] at ($(fp10)-(0,0.55)$) {};
	\node(fp117)[feature] at ($(fp10)-(0,-0.55)$) {};
	\node(fp118)[feature] at ($(fp10)-(-0.275,0.55)$) {};
	\node(fp119)[feature] at ($(fp10)-(-0.275,-0.55)$) {};
	\node(fp120)[feature] at ($(fp10)-(-0.55,0.55)$) {};
	\node(fp121)[feature] at ($(fp10)-(-0.55,0.275)$) {};
	\node(fp122)[feature] at ($(fp10)-(-0.55,0)$) {};
	\node(fp123)[feature] at ($(fp10)-(-0.55,-0.275)$) {};
	\node(fp124)[feature] at ($(fp10)-(-0.55,-0.55)$) {};
	\filldraw[fill=gray_so,rounded corners,fill opacity=0.1,style=dashed] ($(fp10)+(-0.9,-0.9)$) rectangle ($(fp10)+(0.9,0.9)$);
	\draw[arrow1_thick] (annc_trained.east) -- ($(fp10)+(-0.9,0)$);
	
	\node(sfp10)[feature_sfp,fill opacity=1] at ($(struct_model_trained)+(4,0)$) {};
	\node(sfp11)[feature_sfp,fill opacity=0.1] at ($(sfp10)-(0.275,0.275)$) {};
	\node(sfp12)[feature_sfp,fill opacity=0.1] at ($(sfp10)-(0.275,0)$) {};
	\node(sfp13)[feature_sfp,fill opacity=0.1] at ($(sfp10)-(0.275,-0.275)$) {};
	\node(sfp14)[feature_sfp,fill opacity=0.4] at ($(sfp10)-(0,0.275)$) {};
	\node(sfp15)[feature_sfp,fill opacity=0.5] at ($(sfp10)-(0,-0.275)$) {};
	\node(sfp16)[feature_sfp,fill opacity=1,fill=white,style=dashed,draw opacity=0.7] at ($(sfp10)-(-0.275,0.275)$) {};
	\node(sfp17)[feature_sfp,fill opacity=0.6] at ($(sfp10)-(-0.275,0)$) {};
	\node(sfp18)[feature_sfp,fill opacity=0.7] at ($(sfp10)-(-0.275,-0.275)$) {};
	\node(sfp19)[feature_sfp,fill opacity=0.7] at ($(sfp10)-(0.55,0.55)$) {};
	\node(sfp110)[feature_sfp,fill opacity=0.7] at ($(sfp10)-(0.55,0.275)$) {};
	\node(sfp111)[feature_sfp,fill opacity=0.2] at ($(sfp10)-(0.55,0)$) {};
	\node(sfp112)[feature_sfp,fill opacity=0.2] at ($(sfp10)-(0.55,-0.275)$) {};
	\node(sfp113)[feature_sfp,fill opacity=0.2] at ($(sfp10)-(0.55,-0.55)$) {};
	\node(sfp114)[feature_sfp,fill opacity=0.1] at ($(sfp10)-(0.275,0.55)$) {};
	\node(sfp115)[feature_sfp,fill opacity=0.1] at ($(sfp10)-(0.275,-0.55)$) {};
	\node(sfp116)[feature_sfp,fill opacity=0.1] at ($(sfp10)-(0,0.55)$) {};
	\node(sfp117)[feature_sfp,fill opacity=0.1] at ($(sfp10)-(0,-0.55)$) {};
	\node(sfp118)[feature_sfp,fill opacity=0.7] at ($(sfp10)-(-0.275,0.55)$) {};
	\node(sfp119)[feature_sfp,fill opacity=0.8] at ($(sfp10)-(-0.275,-0.55)$) {};
	\node(sfp120)[feature_sfp,fill opacity=0.8] at ($(sfp10)-(-0.55,0.55)$) {};
	\node(sfp121)[feature_sfp,fill opacity=0.7] at ($(sfp10)-(-0.55,0.275)$) {};
	\node(sfp122)[feature_sfp,fill opacity=0.1] at ($(sfp10)-(-0.55,0)$) {};
	\node(sfp123)[feature_sfp,fill opacity=0.1] at ($(sfp10)-(-0.55,-0.275)$) {};
	\node(sfp124)[feature_sfp,fill opacity=0.1] at ($(sfp10)-(-0.55,-0.55)$) {};
	\filldraw[fill=gray_so,rounded corners,fill opacity=0.1,style=dashed] ($(sfp10)+(-0.9,-0.9)$) rectangle ($(sfp10)+(0.9,0.9)$);
	\draw[arrow1_thick] (struct_model_trained.east) -- ($(sfp10)+(-0.9,0)$);
	\draw[line width=3pt] ($(sfp10)+(0,-0.9)$) -- ($(fp10)+(0,0.9)$);
	\coordinate(arrowstart) at ($(fp10)+(0,2.5)$);
	\node(ssfp)[feature,fill=magenta_so,fill opacity=0.7] at ($(arrowstart)+(3,0)$) {N};
	\filldraw[fill=gray_so,rounded corners,fill opacity=0.1,style=dashed] ($(ssfp)+(-0.5,-0.5)$) rectangle ($(ssfp)+(0.5,0.5)$);
	\draw[arrow1_thick] (arrowstart) -- ($(ssfp)-(0.5,0)$);
	\node(conv) at ($(arrowstart)!0.5!($(ssfp)-(0.5,0)$)+(0,0.2)$) {\large fuse};
	\coordinate(centers_new) at ($(ssfp)+(0,-2)$);
	\node(c1_new) at ($(centers_new)+(-0.75,0)$)[feature,fill=magenta_so,text opacity=0]{N};
	\node(center_1_new)[align=center] at ($(c1_new.south)-(0,0.3)$) {$\widehat{c}_1$};
	\node(c2_new) at ($(c1_new)+(0.5,0)$)[feature,fill=magenta_so,text opacity=0]{N};
	\node(center_2_new)[align=center] at ($(c2_new.south)-(0,0.3)$) {$\widehat{c}_2$};
	\node(c3_new) at ($(c2_new)+(0.5,0)$)[feature,fill=magenta_so,text opacity=0,fill opacity=0.1,style=dashed, draw opacity=0.7]{N};
	\node(center_3_new)[align=center] at ($(c3_new.south)-(0,0.3)$) {$\cdots$};
	\node(c4_new) at ($(c3_new)+(0.5,0)$)[feature,fill=magenta_so,text opacity=0]{N};
	\node(center_K_new)[align=center] at ($(c4_new.south)-(0,0.3)$) {$\widehat{c}_K$};
	\filldraw[fill=gray_so,rounded corners,fill opacity=0.1,style=dashed] ($(c1_new)+(-0.5,-0.8)$) rectangle ($(c4_new)+(0.5,0.5)$);
	\node at ($(centers_new)+(2.4,-0.6)$) [rounded corners,style=dashed,color=gray_so] {Class Centers};
	\draw[arrow2,line width=2pt] (annc_pretrain.south) to ++(0,-0.7) to ++(11,0) to ($(centers_new)+(0,-0.8)$);
	\draw[line width=3pt] ($(ssfp)+(0,-0.5)$) -- ($(centers_new)+(0,0.5)$);
	\node(label)[format_rect,fill=red_so,fill opacity=0.3,text opacity=1] at ($($($(ssfp)+(0,-0.5)$)!0.5!($(centers_new)+(0,0.5)$)$)+(2.5,0)$) {Predicted\\ label};
	\draw[arrow1_thick] ($($(ssfp)+(0,-0.5)$)!0.5!($(centers_new)+(0,0.5)$)$) -- (label.west);
	\endscope
	\end{tikzpicture}	
	\caption{\label{flowchart} Flowchart of the proposed framework. The three components are represented by three dashed
		rectangles. The training and testing stages are distinguished by thin and thick solid arrows, respectively.
		The dashed arrows represent the estimation of class centers of features. The intersection
		of neighborhood and training samples are represented by the white pixel, and are excluded at the testing stage.}
}
\end{figure*}

As discussed previously, training with pixel patches may further aggregate the lack of labeled samples.
Based on this concern, the training stage of the proposed framework is performed merely using
the spectral data, while the spatial information is involved posteriorly at the testing stage.
The proposed framework is designed to consist of three parts, namely
\begin{itemize}
	\item  spectral feature extraction;
	\item  spatial structure extraction and fusion of spectral feature and spatial structure;
	\item  classification of the spectral-spatial feature.
\end{itemize}
The flowchart of the whole framework is given in \cref{flowchart}.

In the first part, we extract the spectral features of the whole hyperspectral image by directly applying the
ANNC model proposed in \cite{guo2017spectral}.
This spectrum-based model is supervised with a joint loss of center loss and softmax loss, producing discriminative spectral
features with inner-class compactness and inter-class variations.
Taking advantage of this, the spectral features obtained by the learned
ANNC model is suitable for the successive spatial information fusion step.
In the following,
we use the notations $x$ and $f_{\mathrm{ANNC}}(x)$ to denote a given pixel and its corresponding spectral
feature extracted by the ANNC model.

In the second part,
a spatial structure extraction algorithm is designed to explore the local structure for a given pixel
within some pre-defined neighborhood. More precisely, the local structure is characterized by a customized
``convolutional kernel''. The fusion of the spectral-spatial information is realized by a ``convolutional''
operation between the kernel and the extracted spectral features within the neighborhood.
Let $N(x)$ be a neighborhood centering at pixel $x$ and $W(N(x))$ be the
customized ``convolution kernel''. Use the notation $f_{\mathrm{ANNC}}(N(x))$
to represent the data cube formed by the extracted spectral features
 within the neighborhood $N(x)$, where the $i,j$-th entry is defined by
 \begin{equation}
 f_{\mathrm{ANNC}}(N(x))_{i,j} = f_{\mathrm{ANNC}}(N(x)_{i,j}).
 \end{equation}
Using $f_{N(x)}(x)$ to denote the spectral-spatial feature of $x$ within the neighborhood $N(x)$, it is defined by
\begin{equation}\label{fusion}
    f_{N(x)}(x) = W(N(x))\bigotimes f_{\mathrm{ANNC}}(N(x)),
\end{equation}
where $\bigotimes$ indicates the  convolutional operation.
A more clear interpretation of this model will be given in  \cref{subsection_discriminative_model}.

The last part of the framework aims to classify the resulting spectral-spatial feature $f_{N(x)}(x)$.
In \cite{guo2017spectral},
a center classifier
is introduced, which assigns the label to each sample according to its nearest class center.
Let $\hat{c}_1,\hat{c}_2,\ldots,\hat{c}_k$
be the class centers estimated from the extracted spectral features of the training set.
The label of the spectral-spatial feature $f_{N(x)}(x)$ is predicted by
\begin{equation}\label{center_classifier}
	\ell(f_{N(x)}(x)) = \argmin_{i}\{||f_{N(x)}(x)-\hat{c}_i||_2\}.
\end{equation}
Experiments in \cref{experiment} will show that the spectral-spatial features are not sensitive to the
applied classifiers, that similar classification accuracies are obtained by using the center classifier,
the support vector machine (SVM) algorithm, and the $k$-nearest neighbors ($k$NN) algorithm.
However, the center classifier is still the simple and straightforward one, and is engaged as the default
classifier in the proposed framework.

\subsection{Spatial structure extraction algorithm}\label{subsection_discriminative_model}
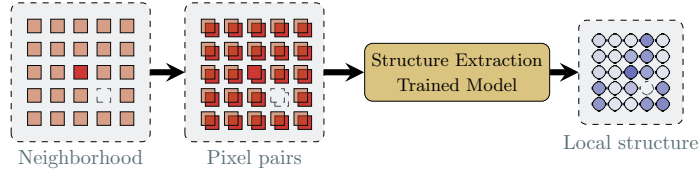
\begin{figure*}
{\linespread{1}
	\centering
	\tikzstyle{format}=[circle,draw,thin,fill=white]
	\tikzstyle{format_gray}=[circle,draw,thin,fill=gray]
	\tikzstyle{format_rect}=[rectangle,draw,thin,fill=white,align=center]
	\tikzstyle{arrowstyle} = [->,thick]
	\tikzstyle{network} = [rectangle, minimum width = 3cm, minimum height = 1cm, text centered, draw = black,align=center,rounded corners,fill=green_so,fill opacity=0.5,text opacity=1]
	\tikzstyle{training_batch} = [trapezium, trapezium left angle = 30, trapezium right angle = 150, minimum width = 3cm, text centered, draw = black, fill = cyan_so, fill opacity=0.3,text opacity=1,align=center]		
	\tikzstyle{class_features} = [trapezium, trapezium left angle = 30, trapezium right angle = 150, minimum width = 3cm, text centered, draw = black, fill = cyan_so, fill opacity=0.3,text opacity=1,align=center]
	\tikzstyle{pixel} = [rectangle, draw = black, fill = orange_so, fill opacity=0.5,text opacity=0,align=center]	
	\tikzstyle{pixel_red} = [rectangle, draw = black, fill = red_so, fill opacity=1,text opacity=0,align=center]	
	\tikzstyle{feature} = [rectangle, draw = black, fill = orange_so, fill opacity=0.3,text opacity=0,align=center,rounded corners]	
	\tikzstyle{feature_sfp} = [rectangle, draw = black, fill = violet_so, fill opacity=0.3,text opacity=0,align=center,rounded corners]					
	\tikzstyle{arrow1} = [thick, ->, >= stealth]
	\tikzstyle{arrow1_thick} = [thick, ->, >= stealth, line width=3pt]
	\tikzstyle{arrow2} = [thick, dashed, ->, >= stealth]
	\begin{tikzpicture}[auto,>=latex',  thin,  start chain=going below, every join/.style={norm}]
	\definecolor{gray_so}{RGB}{88,110,117}
	\definecolor{yellow_so}{RGB}{181,137,0}
	\definecolor{cyan_so}{RGB}{42,161,152}
	\definecolor{orange_so}{RGB}{203,75,22}
	\definecolor{green_so}{RGB}{133,153,0}
	\definecolor{red_so}{RGB}{220,50,47}
	\definecolor{magenta_so}{RGB}{211,54,130}
	\definecolor{violet_so}{RGB}{108,113,196}
	\useasboundingbox (0,0) rectangle (12.7*0.77,3.7*0.77);
	\scope[transform canvas={scale=0.77}]
	
	\coordinate (zero_1) at (0,0);
	
	\node(p10)[pixel_red] at ($(zero_1)+(1.7,2)$) {};
	\node(p11)[pixel] at ($(p10)-(0.4,0.4)$) {};
	\node(p12)[pixel] at ($(p10)-(0.4,0)$) {};
	\node(p13)[pixel] at ($(p10)-(0.4,-0.4)$) {};
	\node(p14)[pixel] at ($(p10)-(0,0.4)$) {};
	\node(p15)[pixel] at ($(p10)-(0,-0.4)$) {};
	\node(p16)[pixel,fill opacity=1,fill=white,style=dashed,draw opacity=0.7] at ($(p10)-(-0.4,0.4)$) {};
	\node(p17)[pixel] at ($(p10)-(-0.4,0)$) {};
	\node(p18)[pixel] at ($(p10)-(-0.4,-0.4)$) {};
	\node(p19)[pixel] at ($(p10)-(0.8,0.8)$) {};
	\node(p110)[pixel] at ($(p10)-(0.8,0.4)$) {};
	\node(p111)[pixel] at ($(p10)-(0.8,0)$) {};
	\node(p112)[pixel] at ($(p10)-(0.8,-0.4)$) {};
	\node(p113)[pixel] at ($(p10)-(0.8,-0.8)$) {};
	\node(p114)[pixel] at ($(p10)-(0.4,0.8)$) {};
	\node(p115)[pixel] at ($(p10)-(0.4,-0.8)$) {};
	\node(p116)[pixel] at ($(p10)-(0,0.8)$) {};
	\node(p117)[pixel] at ($(p10)-(0,-0.8)$) {};
	\node(p118)[pixel] at ($(p10)-(-0.4,0.8)$) {};
	\node(p119)[pixel] at ($(p10)-(-0.4,-0.8)$) {};
	\node(p120)[pixel] at ($(p10)-(-0.8,0.8)$) {};
	\node(p121)[pixel] at ($(p10)-(-0.8,0.4)$) {};
	\node(p122)[pixel] at ($(p10)-(-0.8,0)$) {};
	\node(p123)[pixel] at ($(p10)-(-0.8,-0.4)$) {};
	\node(p124)[pixel] at ($(p10)-(-0.8,-0.8)$) {};
	\filldraw[fill=gray_so,rounded corners,fill opacity=0.1,style=dashed] ($(p10)+(-1.2,-1.2)$) rectangle ($(p10)+(1.2,1.2)$);
	\node at ($(p10)+(0,-1.5)$) [rounded corners,style=dashed,color=gray_so] {Neighborhood};
	
	\node(pwp_10)[pixel_red] at ($(p10)+(3.07,-0.07)$) {};
	\node(pwp_11)[pixel_red] at ($(pwp_10)-(0.4,0.4)$) {};
	\node(pwp_12)[pixel_red] at ($(pwp_10)-(0.4,0)$) {};
	\node(pwp_13)[pixel_red] at ($(pwp_10)-(0.4,-0.4)$) {};
	\node(pwp_14)[pixel_red] at ($(pwp_10)-(0,0.4)$) {};
	\node(pwp_15)[pixel_red] at ($(pwp_10)-(0,-0.4)$) {};
	\node(pwp_16)[pixel, fill opacity=1,fill=white,style=dashed,draw opacity=0.7] at ($(pwp_10)-(-0.4,0.4)$) {};
	\node(pwp_17)[pixel_red] at ($(pwp_10)-(-0.4,0)$) {};
	\node(pwp_18)[pixel_red] at ($(pwp_10)-(-0.4,-0.4)$) {};
	\node(pwp_19)[pixel_red] at ($(pwp_10)-(0.8,0.8)$) {};
	\node(pwp_110)[pixel_red] at ($(pwp_10)-(0.8,0.4)$) {};
	\node(pwp_111)[pixel_red] at ($(pwp_10)-(0.8,0)$) {};
	\node(pwp_112)[pixel_red] at ($(pwp_10)-(0.8,-0.4)$) {};
	\node(pwp_113)[pixel_red] at ($(pwp_10)-(0.8,-0.8)$) {};
	\node(pwp_114)[pixel_red] at ($(pwp_10)-(0.4,0.8)$) {};
	\node(pwp_115)[pixel_red] at ($(pwp_10)-(0.4,-0.8)$) {};
	\node(pwp_116)[pixel_red] at ($(pwp_10)-(0,0.8)$) {};
	\node(pwp_117)[pixel_red] at ($(pwp_10)-(0,-0.8)$) {};
	\node(pwp_118)[pixel_red] at ($(pwp_10)-(-0.4,0.8)$) {};
	\node(pwp_119)[pixel_red] at ($(pwp_10)-(-0.4,-0.8)$) {};
	\node(pwp_120)[pixel_red] at ($(pwp_10)-(-0.8,0.8)$) {};
	\node(pwp_121)[pixel_red] at ($(pwp_10)-(-0.8,0.4)$) {};
	\node(pwp_122)[pixel_red] at ($(pwp_10)-(-0.8,0)$) {};
	\node(pwp_123)[pixel_red] at ($(pwp_10)-(-0.8,-0.4)$) {};
	\node(pwp_124)[pixel_red] at ($(pwp_10)-(-0.8,-0.8)$) {};
	
	\node(pwp10)[pixel_red] at ($(pwp_10)+(-0.07,0.07)$) {};
	\node(pwp11)[pixel] at ($(pwp10)-(0.4,0.4)$) {};
	\node(pwp12)[pixel] at ($(pwp10)-(0.4,0)$) {};
	\node(pwp13)[pixel] at ($(pwp10)-(0.4,-0.4)$) {};
	\node(pwp14)[pixel] at ($(pwp10)-(0,0.4)$) {};
	\node(pwp15)[pixel] at ($(pwp10)-(0,-0.4)$) {};
	\node(pwp16)[pixel,fill opacity=1,fill=white,style=dashed,draw opacity=0.7] at ($(pwp10)-(-0.4,0.4)$) {};
	\node(pwp17)[pixel] at ($(pwp10)-(-0.4,0)$) {};
	\node(pwp18)[pixel] at ($(pwp10)-(-0.4,-0.4)$) {};
	\node(pwp19)[pixel] at ($(pwp10)-(0.8,0.8)$) {};
	\node(pwp110)[pixel] at ($(pwp10)-(0.8,0.4)$) {};
	\node(pwp111)[pixel] at ($(pwp10)-(0.8,0)$) {};
	\node(pwp112)[pixel] at ($(pwp10)-(0.8,-0.4)$) {};
	\node(pwp113)[pixel] at ($(pwp10)-(0.8,-0.8)$) {};
	\node(pwp114)[pixel] at ($(pwp10)-(0.4,0.8)$) {};
	\node(pwp115)[pixel] at ($(pwp10)-(0.4,-0.8)$) {};
	\node(pwp116)[pixel] at ($(pwp10)-(0,0.8)$) {};
	\node(pwp117)[pixel] at ($(pwp10)-(0,-0.8)$) {};
	\node(pwp118)[pixel] at ($(pwp10)-(-0.4,0.8)$) {};
	\node(pwp119)[pixel] at ($(pwp10)-(-0.4,-0.8)$) {};
	\node(pwp120)[pixel] at ($(pwp10)-(-0.8,0.8)$) {};
	\node(pwp121)[pixel] at ($(pwp10)-(-0.8,0.4)$) {};
	\node(pwp122)[pixel] at ($(pwp10)-(-0.8,0)$) {};
	\node(pwp123)[pixel] at ($(pwp10)-(-0.8,-0.4)$) {};
	\node(pwp124)[pixel] at ($(pwp10)-(-0.8,-0.8)$) {};
	
	\filldraw[fill=gray_so,rounded corners,fill opacity=0.1,style=dashed] ($(pwp10)+(-1.2,-1.2)$) rectangle ($(pwp10)+(1.2,1.2)$);
	\node at ($(pwp10)+(0,-1.5)$) [rounded corners,style=dashed,color=gray_so] {Pixel pairs};
	
	\node(struct_model_trained) at ($(pwp10)+(3.5,0)$)[network,fill=yellow_so] {\small{Structure Extraction}\\ \small{Trained Model}};
	
	\node(sfp10)[feature_sfp,fill opacity=1] at ($(struct_model_trained)+(3,0)$) {};
	\node(sfp11)[feature_sfp,fill opacity=0.1] at ($(sfp10)-(0.275,0.275)$) {};
	\node(sfp12)[feature_sfp,fill opacity=0.1] at ($(sfp10)-(0.275,0)$) {};
	\node(sfp13)[feature_sfp,fill opacity=0.1] at ($(sfp10)-(0.275,-0.275)$) {};
	\node(sfp14)[feature_sfp,fill opacity=0.4] at ($(sfp10)-(0,0.275)$) {};
	\node(sfp15)[feature_sfp,fill opacity=0.5] at ($(sfp10)-(0,-0.275)$) {};
	\node(sfp16)[feature_sfp,fill opacity=1,fill=white,style=dashed,draw opacity=0.7] at ($(sfp10)-(-0.275,0.275)$) {};
	\node(sfp17)[feature_sfp,fill opacity=0.6] at ($(sfp10)-(-0.275,0)$) {};
	\node(sfp18)[feature_sfp,fill opacity=0.7] at ($(sfp10)-(-0.275,-0.275)$) {};
	\node(sfp19)[feature_sfp,fill opacity=0.7] at ($(sfp10)-(0.55,0.55)$) {};
	\node(sfp110)[feature_sfp,fill opacity=0.7] at ($(sfp10)-(0.55,0.275)$) {};
	\node(sfp111)[feature_sfp,fill opacity=0.2] at ($(sfp10)-(0.55,0)$) {};
	\node(sfp112)[feature_sfp,fill opacity=0.2] at ($(sfp10)-(0.55,-0.275)$) {};
	\node(sfp113)[feature_sfp,fill opacity=0.2] at ($(sfp10)-(0.55,-0.55)$) {};
	\node(sfp114)[feature_sfp,fill opacity=0.1] at ($(sfp10)-(0.275,0.55)$) {};
	\node(sfp115)[feature_sfp,fill opacity=0.1] at ($(sfp10)-(0.275,-0.55)$) {};
	\node(sfp116)[feature_sfp,fill opacity=0.1] at ($(sfp10)-(0,0.55)$) {};
	\node(sfp117)[feature_sfp,fill opacity=0.1] at ($(sfp10)-(0,-0.55)$) {};
	\node(sfp118)[feature_sfp,fill opacity=0.7] at ($(sfp10)-(-0.275,0.55)$) {};
	\node(sfp119)[feature_sfp,fill opacity=0.8] at ($(sfp10)-(-0.275,-0.55)$) {};
	\node(sfp120)[feature_sfp,fill opacity=0.8] at ($(sfp10)-(-0.55,0.55)$) {};
	\node(sfp121)[feature_sfp,fill opacity=0.7] at ($(sfp10)-(-0.55,0.275)$) {};
	\node(sfp122)[feature_sfp,fill opacity=0.1] at ($(sfp10)-(-0.55,0)$) {};
	\node(sfp123)[feature_sfp,fill opacity=0.1] at ($(sfp10)-(-0.55,-0.275)$) {};
	\node(sfp124)[feature_sfp,fill opacity=0.1] at ($(sfp10)-(-0.55,-0.55)$) {};
	\filldraw[fill=gray_so,rounded corners,fill opacity=0.1,style=dashed] ($(sfp10)+(-0.9,-0.9)$) rectangle ($(sfp10)+(0.9,0.9)$);
	\node at ($(sfp10)+(0,-1.2)$) [rounded corners,style=dashed,color=gray_so] {Local structure};
	
	\draw[arrow1_thick] ($(p10)+(1.2,0)$) -- ($(pwp10)-(1.2,0)$);
	\draw[arrow1_thick] ($(pwp10)+(1.2,0)$) -- (struct_model_trained.west);
	\draw[arrow1_thick] (struct_model_trained.east) -- ($(sfp10)+(-0.9,0)$);
	\endscope
	\end{tikzpicture}	
	\caption{\label{flowchart_sse} Flowchart of the spatial structure extraction algorithm. Pixel pairs are firstly generated by the center pixel and its neighbors
		and then fed into the model. For each pixel-pair, the model predicts the probability that the two elements share the same class label.
		The spatial structure is extracted as the output of this discriminant model.}
}
\end{figure*}

The most crucial aspect of CSFF is the spatial structure extraction algorithm. This algorithm
relies on a CNN-based discriminant network, which
is designed to predict the probability that an input pair of pixels have the same label.
By applying this model to the pixel-pairs generated by the test pixel and its neighbors, the corresponding
local spatial structure can be extracted and represented as a matrix that maps the neighborhood.
The procedure is illustrated in \cref{flowchart_sse}. To be more precise,
use $D((x,x'))$ to denote the predicted probability that the entries of pixel-pair
$(x,x')$ have the same class label. Let $\{x\}\times N(x)=\{(x,x')|\, \forall x'\in N(x)\}$ be a set
of pixel pairs generated by the centering pixel $x$ and its neighbors within $N(x)$, we use
$D(\{x\}\times N(x))$ to denote the corresponding spatial matrix, whose $i,j$-th entry is defined by
\begin{equation}\label{eq:matrix}
D(\{x\}\times N(x))_{i,j} = D((x,N(x)_{i,j})).
\end{equation}
By applying an element-wise threshold function $f_t$, the real-valued $D(\{x\}\times N(x))$
is calibrated to a binary matrix
\begin{equation}\label{eq:binarymatrix}
D_t(\{x\}\times N(x)) = f_t(D(\{x\}\times N(x))),
\end{equation}
where $t$ is the pre-determined threshold value.
The convolutional kernel $W(N(x))$ in\cref{fusion} is calculated by normalizing $D_{t}(\{x\}\times N(x))$ such that all the
elements add up to a unit. In essence, the convolutional kernel $W(N(x))$ records the positions where the spectra are supposed to have the same class label as the centering test pixel. Thus, only the neighboring pixels which correspond to the non-zero positions in $W(N(x))$ will contribute to the estimation of spectral-spatial feature in\cref{fusion}, in a equal manner.
In practice, both the size of neighborhood $N(x)$ and the threshold value $t$ should be appropriately chosen, as to be analysed in \cref{experiment}.

To enhance the hierarchical representative capability of the whole framework,
the aforementioned discriminant model is designed to extract the deep features, while the
spectral feature extraction model (ANNC) focuses on exploring the shallow features with a simple network structure.
From this point of view, a diverse structure from ANNC should be considered, in order to make the proposed discriminant model
different from ANNC, namely
\begin{enumerate}
	\item While ANNC uses a structure of fully connected layers, the CNN architecture will be employed in the
	discriminant model.
	A CNN-based model is not only built differently from ANNC model,
	but is also expected to be more powerful in expressing hierarchical features.
	\item While ANNC uses a shallow structure of $3$ layers, a deep model with more layers will be engaged
	in the discriminant model.
	In most cases, deep networks enable the extraction of
	more abstract features than the respectively shallow ones.
\end{enumerate}

It is noteworthy that training a complex discriminant model with a limited number of labeled pixels is
possible. To address the task of distinguishing whether two spectra belong to the same class,
the discriminant model is designed to be fed with pixel-pairs at
both training and testing stages. Without considering the geometric information,
the pixel-pairs for training are generated by the pixels chosen from training data.
According to this settlement, the training set can be enlarged to
roughly have a squared quantity of training samples.
In practice, the enlarged training set is sufficient
for training our discriminant model.

Based on above discussions, we choose to modify the pixel-pair features (PPF) model
proposed in \cite{li2017hyperspectral} and use the resulting network as the discriminant model in our
spatial structure extraction algorithm. The PPF is a CNN-based framework proposed for multi-classification
task.
The authors designed this model for classifying pixel-pairs into $K+1$ classes, which are
\begin{itemize}
	\item class $i$, if two pairing pixels are from the same class $i$, for $1\le i\le K$;
	\item class $0$, otherwise.
	
\end{itemize}
The pixel-pair strategy is applied for alleviating the shortage
of training data.

Fortunately, the PPF model is not only CNN-based, but also has a relatively deep structure with $9$
convolutional or fully connected layers and $3$ pooling layers.
This meets the directions discussed above for structure design of our discriminant model.
Moreover, the binary classification task in our problem is relatively easier than the original multi-classification task in PPF,
that should be well-tackled by a similar model with PPF.
By modifying the first data layer and the last two fully connected layers, the structure is able to
address different datasets and is suitable for the binary classification task. The modified structure of PPF model
is applied as our discriminant model, which is a part of the spatial feature extraction algorithm.

\begin{figure*}
{\linespread{1}
	\centering
	\tikzstyle{format}=[circle,draw,thin,fill=white]
	\tikzstyle{format_gray}=[circle,draw,thin,fill=gray]
	\tikzstyle{format_rect}=[rectangle,draw,thin,fill=white,align=center]
	\tikzstyle{arrowstyle} = [->,thick]
	\tikzstyle{network} = [rectangle, minimum width = 3cm, minimum height = 1cm, text centered, draw = black,align=center,rounded corners,fill=green_so,fill opacity=0.5,text opacity=1]
	\tikzstyle{training_batch} = [trapezium, trapezium left angle = 30, trapezium right angle = 150, minimum width = 3cm, text centered, draw = black, fill = cyan_so, fill opacity=0.3,text opacity=1,align=center]		
	\tikzstyle{class_features} = [trapezium, trapezium left angle = 30, trapezium right angle = 150, minimum width = 3cm, text centered, draw = black, fill = cyan_so, fill opacity=0.3,text opacity=1,align=center]
	\tikzstyle{pixel} = [rectangle, draw = black, fill = orange_so, fill opacity=0.5,text opacity=0,align=center]	
	\tikzstyle{pixel_red} = [rectangle, draw = black, fill = red_so, fill opacity=1,text opacity=0,align=center]	
	\tikzstyle{feature} = [rectangle, draw = black, fill = orange_so, fill opacity=0.3,text opacity=0,align=center,rounded corners]	
	\tikzstyle{feature_sfp} = [rectangle, draw = black, fill = violet_so, fill opacity=0.3,text opacity=0,align=center,rounded corners]					
	\tikzstyle{arrow1} = [thick, ->, >= stealth]
	\tikzstyle{arrow1_thick} = [thick, ->, >= stealth, line width=2pt]
	\tikzstyle{arrow2} = [thick, dashed, ->, >= stealth]
	\tikzstyle{thick_line} = [line width=0.7pt,dashed]
	\tikzstyle{channel} = [fill = lightgray_so, fill opacity = 0.7]
	\tikzstyle{channel_shadow} = [fill = gray_so, fill opacity = 0.1, rounded corners]
	\tikzstyle{channel_selected} = [fill = violet_so, fill opacity = 0.7]
	\begin{tikzpicture}[auto,>=latex',  thin,  start chain=going below, every join/.style={norm}]
	\definecolor{gray_so}{RGB}{88,110,117}
	\definecolor{lightgray_so}{RGB}{207,221,221}
	\definecolor{yellow_so}{RGB}{181,137,0}
	\definecolor{cyan_so}{RGB}{42,161,152}
	\definecolor{orange_so}{RGB}{203,75,22}
	\definecolor{green_so}{RGB}{133,153,0}
	\definecolor{red_so}{RGB}{220,50,47}
	\definecolor{magenta_so}{RGB}{211,54,130}
	\definecolor{violet_so}{RGB}{108,113,196}
	\useasboundingbox (0,0) rectangle (18*0.77,8*0.77);

	\scope[transform canvas={scale=0.77}]
	
	\coordinate (zero) at (0,0);
	
	\coordinate (input) at ($(zero)+(0.5,4.5)$);
	\node at ($(input)-(-0.25,0.35)$) {Input};
	\fill[channel_shadow,fill opacity=0.3] ($(input)$) -- ($(input)+(0.5,0)$) -- ($(input)+(2,1)$) -- ($(input)+(1.5,1)$) -- ($(input)$);
	\filldraw[channel] ($(input)$) rectangle ($(input)+(0.25,3)$);
	\filldraw[channel] ($(input)+(0.25,0)$) rectangle ($(input)+(0.5,3)$);
	
	\coordinate (c1) at ($(input)+(2.5,0)$);
	\node at($(c1)-(-0.25,0.35)$) {\small{Channels:} $10$};

	\foreach \x/\xtext in {9,...,0}
	{
		\fill[channel_shadow] ($(c1)+(\x/20,\x/20)$) -- ($(c1)+(0.5,0)+(\x/20,\x/20)$) -- ($(c1)+(1.85,0.9)+(\x/20,\x/20)$) -- ($(c1)+(1.35,0.9)+(\x/20,\x/20)$) -- ($(c1)+(\x/20,\x/20)$);
	}
	\foreach \x/\xtext in {9,...,0}
	{
		\pgfmathparse{Mod(\x,2)==0?1:0}
		\ifnum\pgfmathresult>0
		\filldraw[channel] ($(c1)+(\x/20,\x/20)$) rectangle ($(c1)+(0.25,2.7)+(\x/20,\x/20)$);
        \filldraw[channel] ($(c1)+(0.25,0)+(\x/20,\x/20)$) rectangle ($(c1)+(0.5,2.7)+(\x/20,\x/20)$);
		\else
		\filldraw[channel,fill=gray_so] ($(c1)+(\x/20,\x/20)$) rectangle ($(c1)+(0.25,2.7)+(\x/20,\x/20)$);
        \filldraw[channel,fill=gray_so] ($(c1)+(0.25,0)+(\x/20,\x/20)$) rectangle ($(c1)+(0.5,2.7)+(\x/20,\x/20)$);
		\fi
	}

    \node (conv1) at ($(input)!0.5!(c1)+(0.25,0.3)$){${\mathrm{conv}}\atop{9\times 1}$};
    \draw [arrow1_thick] ($(input)+(0.6,0)$) -- ($(c1)+(-0.1,0)$);
    \filldraw[channel_selected] ($(input)+(0.25,2.3)$) rectangle ($(input)+(0.5,3)$);
    \draw[thick_line] ($(input)+(0.5,3)$) -- ($(c1)+(0.25,2.7)$);
    \draw[thick_line] ($(input)+(0.5,2.3)$)--($(c1)+(0.25,2.45)$);
    \filldraw[channel_selected,fill=cyan_so] ($(c1)+(0.25,2.45)$) rectangle ($(c1)+(0.5,2.7)$);

    \coordinate (c2) at ($(c1)+(2.5,0)$);
    \node at($(c2)-(-0.25,0.35)$) {\small{Channels:} $10$};
    \foreach \x/\xtext in {9,...,0}
    {
    	\fill[channel_shadow] ($(c2)+(\x/20,\x/20)$) -- ($(c2)+(0.25,0)+(\x/20,\x/20)$) -- ($(c2)+(1.6,0.9)+(\x/20,\x/20)$) -- ($(c2)+(1.35,0.9)+(\x/20,\x/20)$) -- ($(c2)+(\x/20,\x/20)$);
    }
    \foreach \x/\xtext in {9,...,0}
    {
    	\pgfmathparse{Mod(\x,2)==0?1:0}
    	\ifnum\pgfmathresult>0
    	\filldraw[channel] ($(c2)+(\x/20,\x/20)$) rectangle ($(c2)+(0.25,2.7)+(\x/20,\x/20)$);
    	\else
    	\filldraw[channel,fill=gray_so] ($(c2)+(\x/20,\x/20)$) rectangle ($(c2)+(0.25,2.7)+(\x/20,\x/20)$);
    	\fi
    }
    \node (conv2) at ($(c1)!0.5!(c2)+(0.25,0.3)$){${\mathrm{conv}}\atop{1\times 2}$};
    \draw [arrow1_thick] ($(c1)+(0.6,0)$) -- ($(c2)+(-0.1,0)$);
    \filldraw[channel_selected] ($(c1)+(0,1.95)$) rectangle ($(c1)+(0.5,2.2)$);
    \draw[thick_line] ($(c1)+(0.5,1.95)$) -- ($(c2)+(0,1.95)$);
    \draw[thick_line] ($(c1)+(0.5,2.2)$)--($(c2)+(0,2.2)$);
    \filldraw[channel_selected,fill=cyan_so] ($(c2)+(0,1.95)$) rectangle ($(c2)+(0.25,2.2)$);

    \coordinate (pool3) at ($(c2)+(2.5,0)$);
    \node at($(pool3)-(-0.25,0.35)$) {\small{Channels:} $20$};
    \foreach \x/\xtext in {9,...,0}
    {
    	\fill[channel_shadow] ($(pool3)+(\x/20,\x/20)$) -- ($(pool3)+(0.25,0)+(\x/20,\x/20)$) -- ($(pool3)+(1.15,0.6)+(\x/20,\x/20)$) -- ($(pool3)+(0.9,0.6)+(\x/20,\x/20)$) -- ($(pool3)+(\x/20,\x/20)$);
    }
	\foreach \x/\xtext in {9,...,0}
	{
		\pgfmathparse{Mod(\x,2)==0?1:0}
		\ifnum\pgfmathresult>0
		\filldraw[channel] ($(pool3)+(\x/20,\x/20)$) rectangle ($(pool3)+(0.25,1.8)+(\x/20,\x/20)$);
		\else
		\filldraw[channel,fill=gray_so] ($(pool3)+(\x/20,\x/20)$) rectangle ($(pool3)+(0.25,1.8)+(\x/20,\x/20)$);
		\fi
	}
    \node (pooling3) at ($(c2)!0.5!(pool3)+(0.125,0.3)$){${\mathrm{pool}}\atop{3\times 1}$};
    \draw [arrow1_thick] ($(c2)+(0.35,0)$) -- ($(pool3)+(-0.1,0)$);
    \filldraw[channel_selected] ($(c2)+(0,0.95)$) rectangle ($(c2)+(0.25,1.7)$);
    \draw[thick_line] ($(c2)+(0.25,0.95)$) -- ($(pool3)+(0,0.7)$);
    \draw[thick_line] ($(c2)+(0.25,1.7)$)--($(pool3)+(0,0.95)$);
    \filldraw[channel_selected,fill=cyan_so] ($(pool3)+(0,0.7)$) rectangle ($(pool3)+(0.25,0.95)$);

    \coordinate (c4) at ($(pool3)+(2.5,0)$);
    \node at($(c4)-(-0.25,0.35)$) {\small{Channels:} $20$};
    \foreach \x/\xtext in {19,...,0}
    {
    	\fill[channel_shadow] ($(c4)+(\x/20,\x/20)$) -- ($(c4)+(0.25,0)+(\x/20,\x/20)$) -- ($(c4)+(1.10,0.57)+(\x/20,\x/20)$) -- ($(c4)+(0.85,0.57)+(\x/20,\x/20)$) -- ($(c4)+(\x/20,\x/20)$);
    }
    \foreach \x/\xtext in {19,...,0}
    {
    	\pgfmathparse{Mod(\x,2)==0?1:0}
    	\ifnum\pgfmathresult>0
    	\filldraw[channel] ($(c4)+(\x/20,\x/20)$) rectangle ($(c4)+(0.25,1.7)+(\x/20,\x/20)$);
    	\else
    	\filldraw[channel,fill=gray_so] ($(c4)+(\x/20,\x/20)$) rectangle ($(c4)+(0.25,1.7)+(\x/20,\x/20)$);
    	\fi
    }
    \node (conv4) at ($(pool3)!0.5!(c4)+(0.125,0.3)$){${\mathrm{conv}}\atop{3\times 1}$};
    \draw [arrow1_thick] ($(pool3)+(0.35,0)$) -- ($(c4)+(-0.1,0)$);
    \filldraw[channel_selected] ($(pool3)+(0,1.05)$) rectangle ($(pool3)+(0.25,1.8)$);
    \draw[thick_line] ($(pool3)+(0.25,1.05)$) -- ($(c4)+(0,1.45)$);
    \draw[thick_line] ($(pool3)+(0.25,1.8)$)--($(c4)+(0,1.7)$);
    \filldraw[channel_selected,fill=cyan_so] ($(c4)+(0,1.45)$) rectangle ($(c4)+(0.25,1.7)$);

    \coordinate (pool5) at ($(c4)+(2.5,0)$);
    \node at($(pool5)-(-0.25,0.35)$) {\small{Channels:} $20$};
    \foreach \x/\xtext in {19,...,0}
    {
    	\fill[channel_shadow] ($(pool5)+(\x/20,\x/20)$) -- ($(pool5)+(0.25,0)+(\x/20,\x/20)$) -- ($(pool5)+(0.8,0.37)+(\x/20,\x/20)$) -- ($(pool5)+(0.55,0.37)+(\x/20,\x/20)$) -- ($(pool5)+(\x/20,\x/20)$);
    }
    \foreach \x/\xtext in {19,...,0}
    {
    	\pgfmathparse{Mod(\x,2)==0?1:0}
    	\ifnum\pgfmathresult>0
    	\filldraw[channel] ($(pool5)+(\x/20,\x/20)$) rectangle ($(pool5)+(0.25,1.1)+(\x/20,\x/20)$);
    	\else
    	\filldraw[channel,fill=gray_so] ($(pool5)+(\x/20,\x/20)$) rectangle ($(pool5)+(0.25,1.1)+(\x/20,\x/20)$);
    	\fi
    }

    \node (pooling5) at ($(c4)!0.5!(pool5)+(0.125,0.3)$){${\mathrm{pool}}\atop{2\times 1}$};
    \draw [arrow1_thick] ($(c4)+(0.35,0)$) -- ($(pool5)+(-0.1,0)$);
    \filldraw[channel_selected] ($(c4)+(0,0.85)$) rectangle ($(c4)+(0.25,1.35)$);
    \draw[thick_line] ($(c4)+(0.25,1.35)$) -- ($(pool5)+(0,0.6)$);
    \draw[thick_line] ($(c4)+(0.25,0.85)$)--($(pool5)+(0,0.35)$);
    \filldraw[channel_selected,fill=cyan_so] ($(pool5)+(0,0.35)$) rectangle ($(pool5)+(0.25,0.6)$);

    \coordinate (c6) at ($(pool5)+(2.5,0)$);
    \node at($(c6)-(-0.25,0.35)$) {\small{Channels:} $40$};
    \foreach \x/\xtext in {39,...,0}
    {
    	\fill[channel_shadow] ($(c6)+(\x/20,\x/20)$) -- ($(c6)+(0.25,0)+(\x/20,\x/20)$) -- ($(c6)+(0.8,0.37)+(\x/20,\x/20)$) -- ($(c6)+(0.55,0.37)+(\x/20,\x/20)$) -- ($(c6)+(\x/20,\x/20)$);
    }
    \foreach \x/\xtext in {39,...,0}
    {
	    \pgfmathparse{Mod(\x,2)==0?1:0}
	    \ifnum\pgfmathresult>0
	    \filldraw[channel] ($(c6)+(\x/20,\x/20)$) rectangle ($(c6)+(0.25,1.1)+(\x/20,\x/20)$);
    	\else
    	\filldraw[channel,fill=gray_so] ($(c6)+(\x/20,\x/20)$) rectangle ($(c6)+(0.25,1.1)+(\x/20,\x/20)$);
    	\fi
    }
    \node (conv6) at ($(pool5)!0.5!(c6)+(0.125,0.3)$){${\mathrm{conv}}\atop{3\times 1}$};
    \draw [arrow1_thick] ($(pool5)+(0.35,0)$) -- ($(c6)+(-0.1,0)$);
    \filldraw[channel_selected] ($(pool5)+(0,0.6)$) rectangle ($(pool5)+(0.25,1.1)$);
    \draw[thick_line] ($(pool5)+(0.25,0.6)$) -- ($(c6)+(0,0.85)$);
    \draw[thick_line] ($(pool5)+(0.25,1.1)$)--($(c6)+(0,1.1)$);
    \filldraw[channel_selected,fill=cyan_so] ($(c6)+(0,0.85)$) rectangle ($(c6)+(0.25,1.1)$);

    \coordinate (c7) at ($(c1)-(0,4)$);
    \node at($(c7)-(-0.25,0.35)$) {\small{Channels:} $40$};
    \foreach \x/\xtext in {39,...,0}
    {
    	\fill[channel_shadow] ($(c7)+(\x/20,\x/20)$) -- ($(c7)+(0.25,0)+(\x/20,\x/20)$) -- ($(c7)+(0.8,0.37)+(\x/20,\x/20)$) -- ($(c7)+(0.55,0.37)+(\x/20,\x/20)$) -- ($(c7)+(\x/20,\x/20)$);
    }
    \foreach \x/\xtext in {39,...,0}
    {
	    \pgfmathparse{Mod(\x,2)==0?1:0}
    	\ifnum\pgfmathresult>0
    	\filldraw[channel] ($(c7)+(\x/20,\x/20)$) rectangle ($(c7)+(0.25,1)+(\x/20,\x/20)$);
    	\else
    	\filldraw[channel,fill=gray_so] ($(c7)+(\x/20,\x/20)$) rectangle ($(c7)+(0.25,1)+(\x/20,\x/20)$);
    	\fi
    }

    \node (conv7) at ($($(input)-(0,4)$)!0.5!(c7)+(0.125,0.3)$){${\mathrm{conv}}\atop{3\times 1}$};
    \draw [arrow1_thick,dashed] ($(c6)+(0.35,0)$) -- ($(c6)+(2.3,0)$);
    \draw [arrow1_thick, dashed] ($(input)-(0,4)+(0.6,0)$) -- ($(c7)+(-0.1,0)$);
    \filldraw[channel_selected] ($(c6)+(0,0)$) rectangle ($(c6)+(0.25,0.5)$);
    \filldraw[channel_selected,fill=cyan_so] ($(c7)+(0,0)$) rectangle ($(c7)+(0.25,0.25)$);

	\coordinate (pool8) at ($(c7)+(2.5,0)$);
	\node at($(pool8)-(-0.25,0.35)$) {\small{Channels:} $40$};
	\foreach \x/\xtext in {39,...,0}
	{
		\fill[channel_shadow] ($(pool8)+(\x/20,\x/20)$) -- ($(pool8)+(0.25,0)+(\x/20,\x/20)$) -- ($(pool8)+(0.6,0.23)+(\x/20,\x/20)$) -- ($(pool8)+(0.35,0.23)+(\x/20,\x/20)$) -- ($(pool8)+(\x/20,\x/20)$);
	}
	\foreach \x/\xtext in {39,...,0}
	{
		\pgfmathparse{Mod(\x,2)==0?1:0}
		\ifnum\pgfmathresult>0
		\filldraw[channel] ($(pool8)+(\x/20,\x/20)$) rectangle ($(pool8)+(0.25,0.7)+(\x/20,\x/20)$);
		\else
		\filldraw[channel,fill=gray_so] ($(pool8)+(\x/20,\x/20)$) rectangle ($(pool8)+(0.25,0.7)+(\x/20,\x/20)$);
		\fi
	}
    \filldraw[channel_selected] ($(c7)+(0,0.5)$) rectangle ($(c7)+(0.25,1)$);
    \draw[thick_line] ($(c7)+(0.25,0.5)$) -- ($(pool8)+(0,0.45)$);
    \draw[thick_line] ($(c7)+(0.25,1)$)--($(pool8)+(0,0.7)$);
    \node (pooling8) at ($(c7)!0.5!(pool8)+(0.125,0.3)+(1,1)-(0.3,0)$){${\mathrm{pool}}\atop{2\times 1}$};
    \draw [arrow1_thick] ($(c7)+(0.35,0)+(1,1)$) -- ($(pool8)+(-0.7,0)+(1,1)$);
    \filldraw[channel_selected,fill=cyan_so] ($(pool8)+(0,0.45)$) rectangle ($(pool8)+(0.25,0.7)$);

	\coordinate (c9) at ($(pool8)+(2.5,0)$);
	\node at($(c9)-(-0.25,0.35)$) {\small{Channels:} $80$};
	\foreach \x/\xtext in {79,...,0}
	{
		\fill[channel_shadow,fill opacity=0.2] ($(c9)+(\x/30,\x/30)$) -- ($(c9)+(0.25,0)+(\x/30,\x/30)$) -- ($(c9)+(0.475,0.15)+(\x/30,\x/30)$) -- ($(c9)+(0.225,0.15)+(\x/30,\x/30)$) -- ($(c9)+(\x/30,\x/30)$);
	}
    \foreach \x/\xtext in {79,...,0}
    {
	    \pgfmathparse{Mod(\x,2)==0?1:0}
	    \ifnum\pgfmathresult>0
	    \filldraw[channel] ($(c9)+(\x/30,\x/30)$) rectangle ($(c9)+(0.25,0.25)+(\x/30,\x/30)$);
	    \else
	    \filldraw[channel,fill=gray_so] ($(c9)+(\x/30,\x/30)$) rectangle ($(c9)+(0.25,0.25)+(\x/30,\x/30)$);
	    \fi
    }
    \filldraw[channel_selected] ($(pool8)+(0,0)$) rectangle ($(pool8)+(0.25,0.7)$);
    \draw[thick_line] ($(pool8)+(0.25,0)$) -- ($(c9)+(0,0)$);
    \draw[thick_line] ($(pool8)+(0.25,0.7)$)--($(c9)+(0,0.25)$);
    \node (conv9) at ($(pool8)!0.5!(c9)+(0.125,0.3)+(1,1)$){${\mathrm{conv}}\atop{\mathrm{height}\times 1}$};
    \draw [arrow1_thick] ($(pool8)+(0.35,0)+(1,1)$) -- ($(c9)+(-0.3,0)+(1,1)$);
    \filldraw[channel_selected,fill=cyan_so] ($(c9)+(0,0)$) rectangle ($(c9)+(0.25,0.25)$);

	\coordinate (f10) at ($(c9)+(2.5,0)$);
	\node at($(f10)-(-0.25,0.35)$) {\small{Channels:} $80$};
    \draw[thick_line] ($(c9)+(0.25,0.125)$) -- ($(f10)+(0,0.125)+(79/30,79/30)$);
    \draw[thick_line] ($(c9)+(0.25,0.125)+(79/30,79/30)$)--($(f10)+(0,0.125)$);
	\foreach \x/\xtext in {79,...,0}
	{
		\fill[channel_shadow,fill opacity=0.2] ($(f10)+(\x/30,\x/30)$) -- ($(f10)+(0.25,0)+(\x/30,\x/30)$) -- ($(f10)+(0.475,0.15)+(\x/30,\x/30)$) -- ($(f10)+(0.225,0.15)+(\x/30,\x/30)$) -- ($(f10)+(\x/30,\x/30)$);
	}
    \foreach \x/\xtext in {79,...,0}
    {
	    \pgfmathparse{Mod(\x,2)==0?1:0}
	    \ifnum\pgfmathresult>0
	    \filldraw[channel] ($(f10)+(\x/30,\x/30)$) rectangle ($(f10)+(0.25,0.25)+(\x/30,\x/30)$);
	    \else
	    \filldraw[channel,fill=gray_so] ($(f10)+(\x/30,\x/30)$) rectangle ($(f10)+(0.25,0.25)+(\x/30,\x/30)$);
	    \fi
    }
    \node (fc10) at ($(c9)!0.5!(f10)+(0.125,0.3)$){FC};
    \draw [arrow1_thick] ($(c9)+(0.35,0)$) -- ($(f10)+(-0.1,0)$);

	\coordinate (f11) at ($(f10)+(2.5,0)$);
	\draw[thick_line] ($(f10)+(0.25,0.125)$) -- ($(f11)+(0,0.25)+(1.2,1.2)$);
	\draw[thick_line] ($(f10)+(0.25,0.125)+(79/30,79/30)$)--($(f11)+(0,0.25)+(1,1)$);
	\foreach \x/\xtext in {1,...,0}
	{
		\fill[channel_shadow,fill opacity=0.2] ($(f11)+(1,1)+(\x/5,\x/5)$) -- ($(f11)+(1,1)+(0.5,0)+(\x/5,\x/5)$) -- ($(f11)+(1,1)+(0.75,0.17)+(\x/5,\x/5)$) -- ($(f11)+(1,1)+(0.25,0.17)+(\x/5,\x/5)$) -- ($(f11)+(1,1)+(\x/5,\x/5)$);
	}
    \foreach \x/\xtext in {1,...,0}
    {
    	\pgfmathparse{Mod(\x,2)==0?1:0}
    	\ifnum\pgfmathresult>0
    	\filldraw[channel] ($(f11)+(1,1)+(\x/5,\x/5)$) rectangle ($(f11)+(1,1)+(0.5,0.5)+(\x/5,\x/5)$);
    	\else
    	\filldraw[channel,fill=gray_so] ($(f11)+(1,1)+(\x/5,\x/5)$) rectangle ($(f11)+(1,1)+(0.5,0.5)+(\x/5,\x/5)$);
    	\fi
    }
	\node at($(f11)+(1,1)-(-0.35,0.35)$) {\small{Channels:} $2$};

    \node (fc11) at ($(f10)!0.5!(f11)+(0.125,0.3)+(1.1,1.1)$){FC};

    \coordinate (softmax12) at ($(f11)+(2.5,0)$);
    \foreach \x/\xtext in {1,...,0}
    {
    	\fill[channel_shadow,fill opacity=0.2] ($(softmax12)+(1,1)+(\x/5,\x/5)$) -- ($(softmax12)+(1,1)+(0.5,0)+(\x/5,\x/5)$) -- ($(softmax12)+(1,1)+(0.75,0.17)+(\x/5,\x/5)$) -- ($(softmax12)+(1,1)+(0.25,0.17)+(\x/5,\x/5)$) -- ($(softmax12)+(1,1)+(\x/5,\x/5)$);
    }
    \foreach \x/\xtext in {1,...,0}
    {
    	\pgfmathparse{Mod(\x,2)==0?1:0}
    	\ifnum\pgfmathresult>0
    	\filldraw[channel] ($(softmax12)+(1,1)+(\x/5,\x/5)$) rectangle ($(softmax12)+(1,1)+(0.5,0.5)+(\x/5,\x/5)$);
    	\else
    	\filldraw[channel,fill=gray_so] ($(softmax12)+(1,1)+(\x/5,\x/5)$) rectangle ($(softmax12)+(1,1)+(0.5,0.5)+(\x/5,\x/5)$);
    	\fi
    }
    \node at ($(softmax12)+(1,1)-(-0.35,0.35)$) {Output};
    \draw [arrow1_thick] ($(f11)+(0.65,0)+(1,1)$) -- ($(softmax12)+(-0.15,0)+(1,1)$);
    \node (fc11) at ($(f11)!0.5!(softmax12)+(0.2,0.22)+(1,1)$){$\mathrm{softmax}$};

	\endscope
	\end{tikzpicture}	
	\caption{\label{dmstructure} Structure of the discriminant model, modified from PPF~\cite{li2017hyperspectral}. The sizes of convolution kernels are marked under
		the layer type $\mathrm{conv}$,
		while the cardinalities of channels are presented under the graphical data blobs.
		In fact, each channel of a preceding data blobs
		is connected to all the channels of latter data blobs in the convolutional layers,
		but only one connection is drawn for simplicity.
		}
}
\end{figure*}
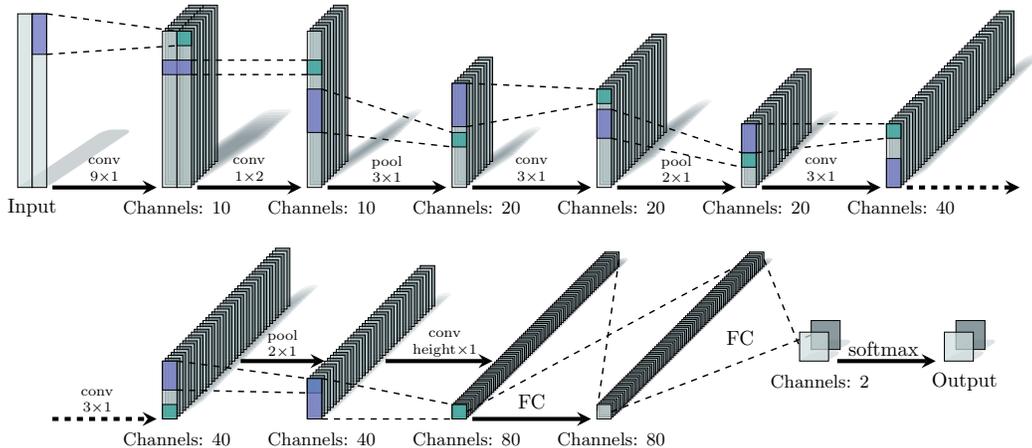

The structure details of the discriminant model are illustrated in \cref{dmstructure}.
Considering the slim shape of the input data, namely $(2,L)$, with $L$ being the number of spectral channels,
the convolution kernels and the pooling regions are chosen as narrow rectangular ones instead of the
traditional squared ones.
All the strides of the convolutional layers are set to be $1$ and the pooling layers are defined to use
the max-pooling function.
In order to introduce nonlinearity to the CNN-based model, the commonly-used rectified linear unit (ReLU)
function, defined by $\phi(x) = \max\{0,x\}$, is applied after every convolutional and
fully connected layer.

\section{Experimental settings and result analysis}\label{sec: Experiments}

\subsection{Datasets description}\label{datasets}
\begin{figure}
	\centering
\graphicspath{{Figures/}}
	\emph{\includegraphics[trim = 25mm 30mm 31mm 25mm, clip,width=.12\textwidth]  {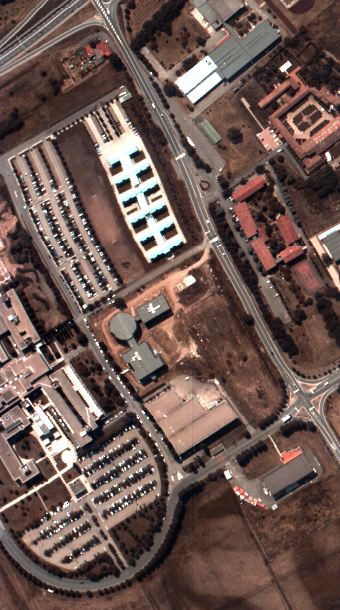}
		\includegraphics[trim = 25mm 30mm 31mm 25mm,clip,width=.12\textwidth]  {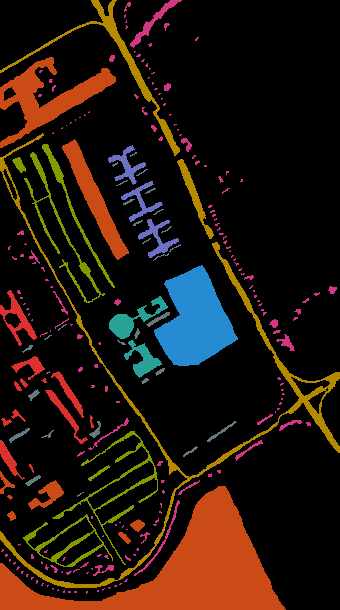}
	}\includegraphics[trim = 0mm 0mm 0mm 0mm,clip,width=.11\textwidth]  {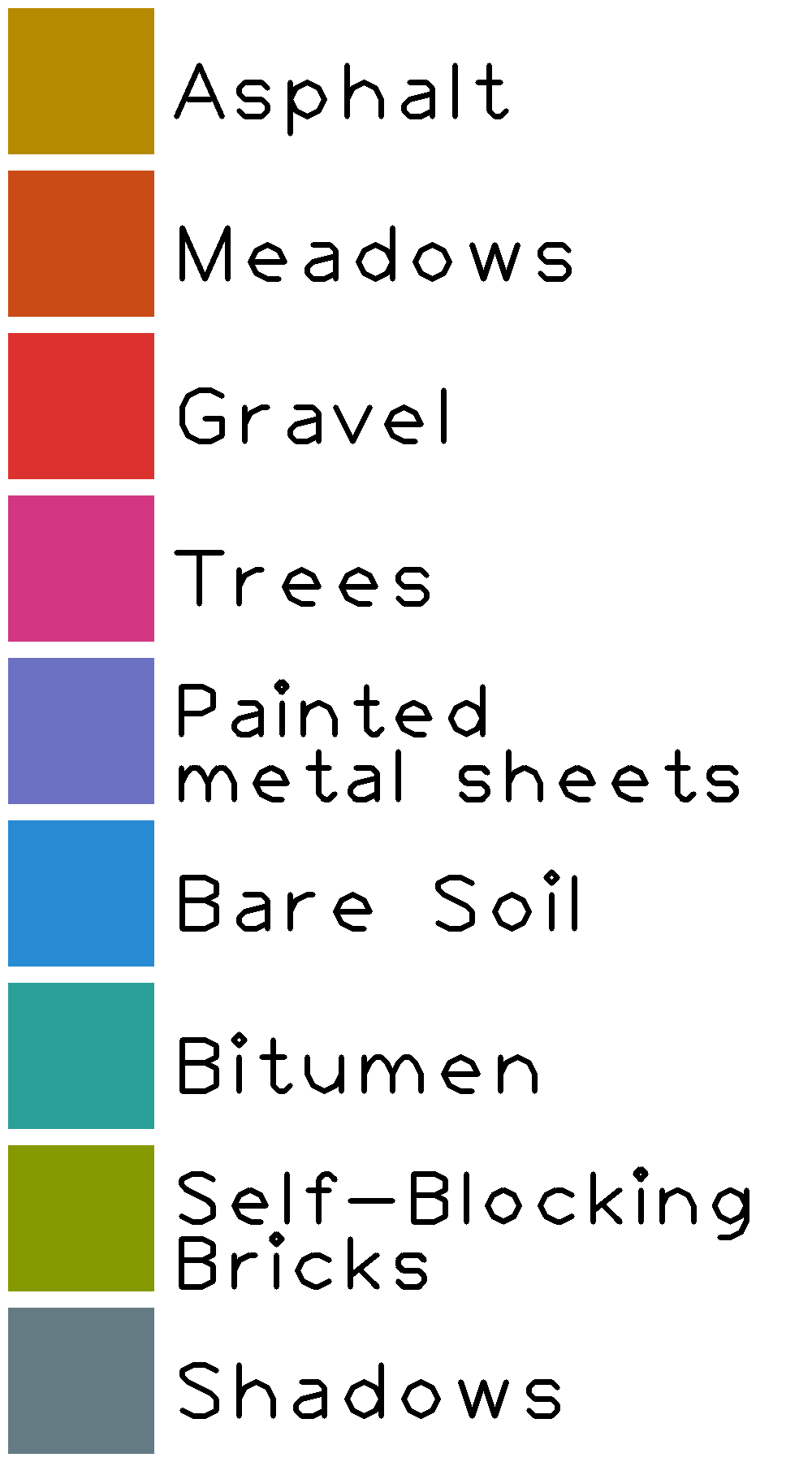}
	\caption{\label{paviau_fc_gt} The false color composite (band $10,20,40$) and groundtruth representation of Pavia University}
\end{figure}

Experiments are performed on three real hyperspectral images, namely the Pavia University scene, the Salinas scene, and the Pavia Centre scene{\footnote{The datasets are available online: \url{http://www.ehu.eus/ccwintco/index.php?title=Hyperspectral_Remote_Sensing_Scenes}}}.
The first data is the Pavia University scene, acquired by the
Reflective Optics System Imaging Spectrometer (ROSIS) sensor. After removing the noisy bands and a blank
strip, a sub-image of $610 \times 340$ pixels with $103$ spectral bands are retained for analysis.
The image is characterized by a spatial resolution of about $1.3$ meters. As summarized in~\cref{label_paviau}, this area is known to be mainly composed by $K=9$ classes of
materials, denoted by labels from $1$ to $9$. The background pixels are represented by label $0$,
and will not be taken into account for classification. \cref{paviau_fc_gt} presents the
false color composite and groundtruth map.
\begin{table}\caption{\label{label_paviau}Reference classes and sizes of training and testing sets of Pavia University image}
	\centering
	\begin{tabular}{c|c|c|c|c}
		\hline
		\hline
		No. 	&Class	&Cardinality	&Train	&Test\\
		\hline
		$1$	&Asphalt	&$6631$	&$200$	&$6431$	\\
		$2$	&Meadows	&$18649$	&$200$	&$18449$	\\
		$3$	&Gravel	&$2099$	&$200$	&$1899$	\\
		$4$	&Trees	&$3064$	&$200$	&$2864$	\\
		$5$	&Painted	metal	sheets	&$1345$	&$200$	&$1145$	\\
		$6$	&Bare	Soil	&$5029$	&$200$	&$4829$	\\
		$7$	&Bitumen	&$1330$	&$200$	&$1130$	\\
		$8$	&Self-Blocking	Bricks	&$3682$	&$200$	&$3482$	\\
		$9$	&Shadows	&$947$	&$200$	&$747$	\\
		\hline
		\multicolumn{2}{c|}{Total}&42776&1800&40976	\\
		\hline
		\hline
	\end{tabular}
\end{table}

\begin{figure}
	\centering
\graphicspath{{Figures/}}
	\includegraphics[trim = 2mm 22mm 13mm 12mm, clip,width=.12\textwidth]  {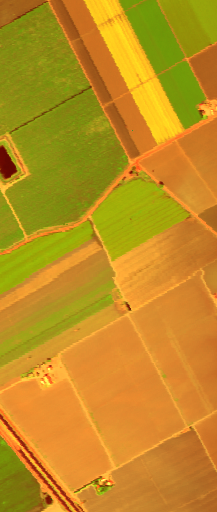}
	\includegraphics[trim =  2mm 22mm 13mm 12mm,clip,width=.12\textwidth]  {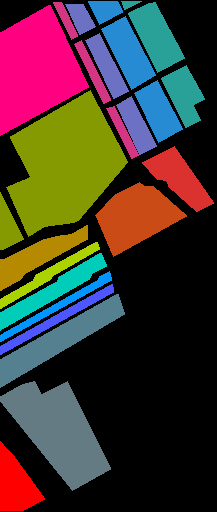}
	\includegraphics[trim =  0mm 0mm 0mm 0mm,clip,width=.1\textwidth]  {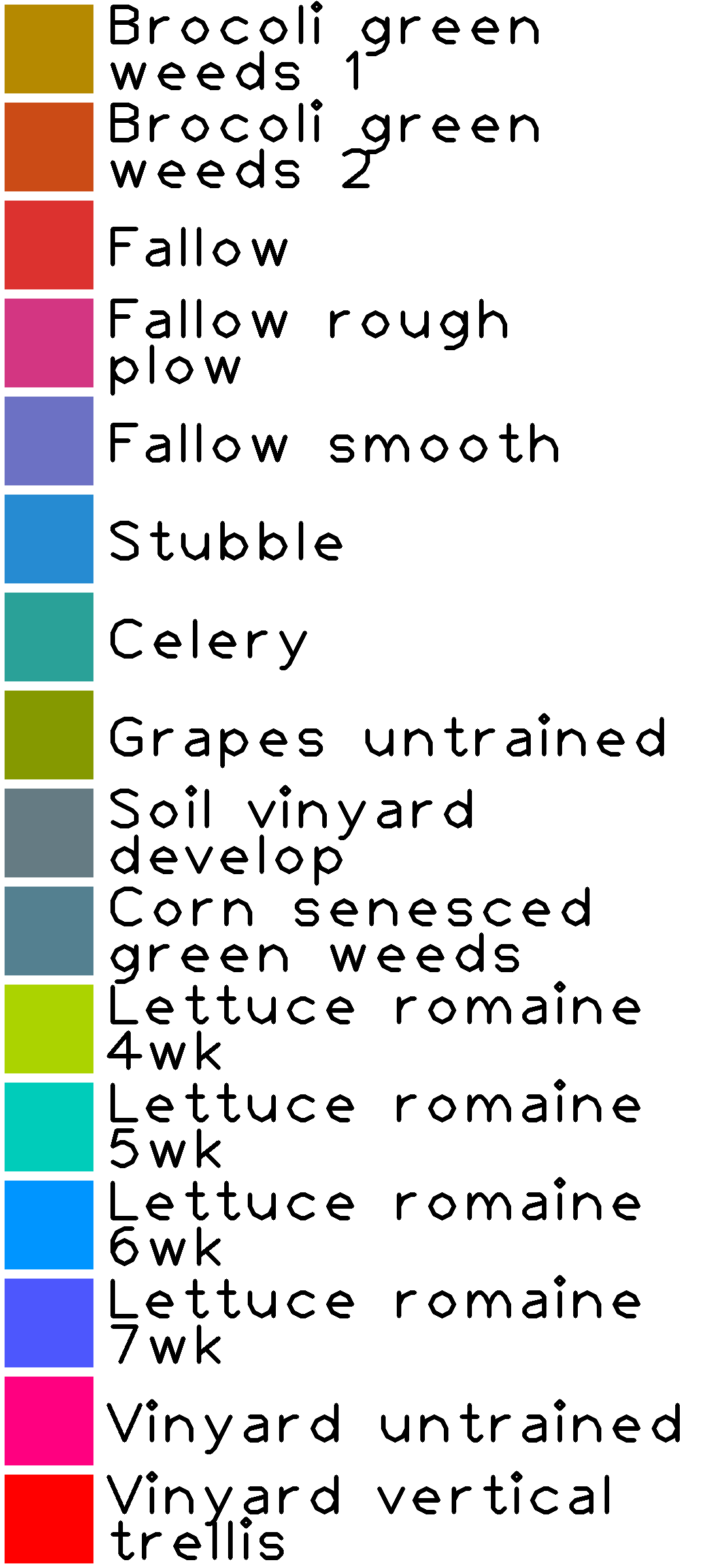}	
	\caption{\label{salinas_fc_gt} The false color composite (band $180,100,10$) and groundtruth representation of Salinas}
\end{figure}

The second one is the Salinas scene collected by the Airborne Visible Infrared Imaging Spectrometer (AVIRIS).
This dataset contains $512\times 217$ pixels,
and is characterized by a resolution of $3.7$ meters. After removing the water absorption bands, the
remaining $204$ (out of $224$) bands are utilized. According to the available groundtruth information
in \cref{label_salinas}, there are $K=16$ composition categories of interest, with the
background pixels represented by label $0$. False color composite and groundtruth map of the Salinas
scene are shown in \cref{salinas_fc_gt}.

\begin{table}\caption{\label{label_salinas}Reference classes and sizes of training and testing sets of Salinas image}
	\centering
	\begin{tabular}{c|c|c|c|c}
		\hline
		\hline
		No. 	&Class	&Cardinality	&Train&   Test	\\
		\hline
		$1$	&Brocoli	green	weeds	1	&$2009$	&$200$	&$1809$	\\
		$2$	&Brocoli	green	weeds	2	&$3726$	&$200$	&$3526$	\\
		$3$	&Fallow	&$1976$	&$200$	&$1776$	\\
		$4$	&Fallow	rough	plow	&$1394$	&$200$	&$1194$	\\
		$5$	&Fallow	smooth	&$2678$	&$200$	&$2478$	\\
		$6$	&Stubble	&$3959$	&$200$	&$3759$	\\
		$7$	&Celery	&$3579$	&$200$	&$3379$	\\
		$8$	&Grapes	untrained	&$11271$	&$200$	&$11071$	\\
		$9$	&Soil	vinyard	develop	&$6203$	&$200$	&$6003$	\\
		$10$	&Corn	senesced	green	weeds	&$3278$	&$200$	&$3078$	\\
		$11$	&Lettuce	romaine	4wk	&$1068$	&$200$	&$868$	\\
		$12$	&Lettuce	romaine	5wk	&$1927$	&$200$	&$1727$	\\
		$13$	&Lettuce	romaine	6wk	&$916$	&$200$	&$716$	\\
		$14$	&Lettuce	romaine	7wk	&$1070$	&$200$	&$870$	\\
		$15$	&Vinyard	untrained	&$7268$	&$200$	&$7068$	\\
		$16$	&Vinyard	vertical	trellis	&$1807$	&$200$	&$1607$	\\
		\hline
		\multicolumn{2}{c|}{Total}&54129&3200&50929	\\
		\hline
		\hline
	\end{tabular}
\end{table}

\begin{figure}[htb]
	\centering
\graphicspath{{Figures/}}
	\includegraphics[trim =16mm 20mm 25mm 10mm, clip,width=.12\textwidth] {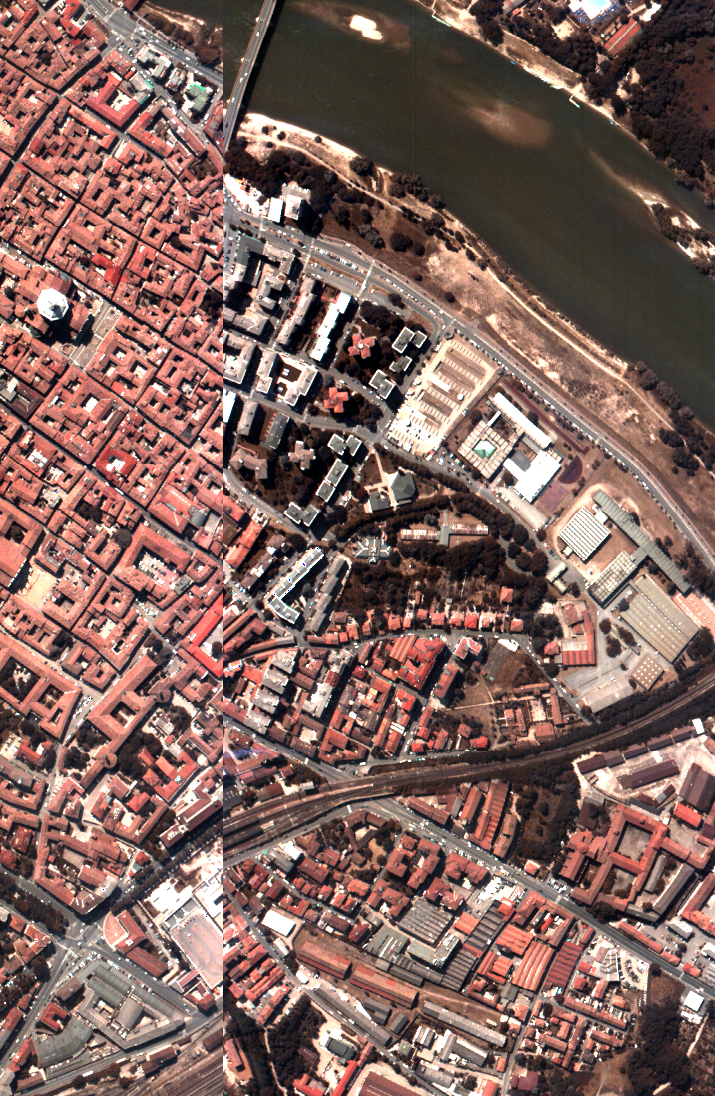}
	\includegraphics[trim =16mm 20mm 25mm 10mm, clip,width=.12\textwidth]{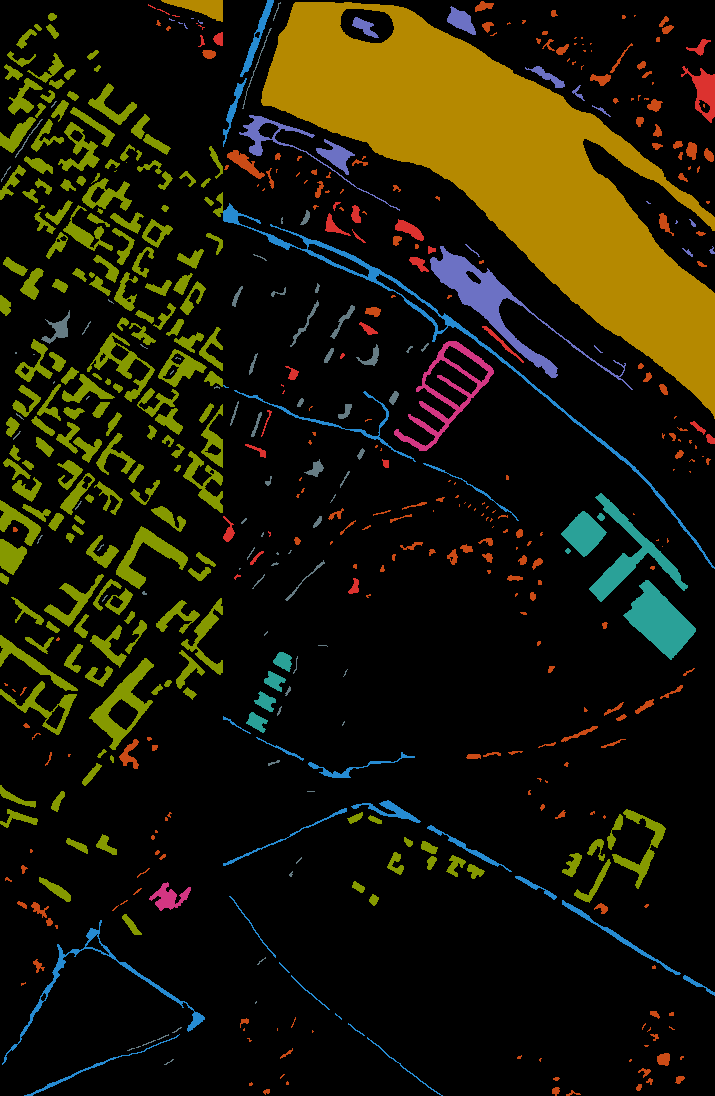}
	\includegraphics[trim =0mm 0mm 0mm 0mm, clip,width=.1\textwidth]{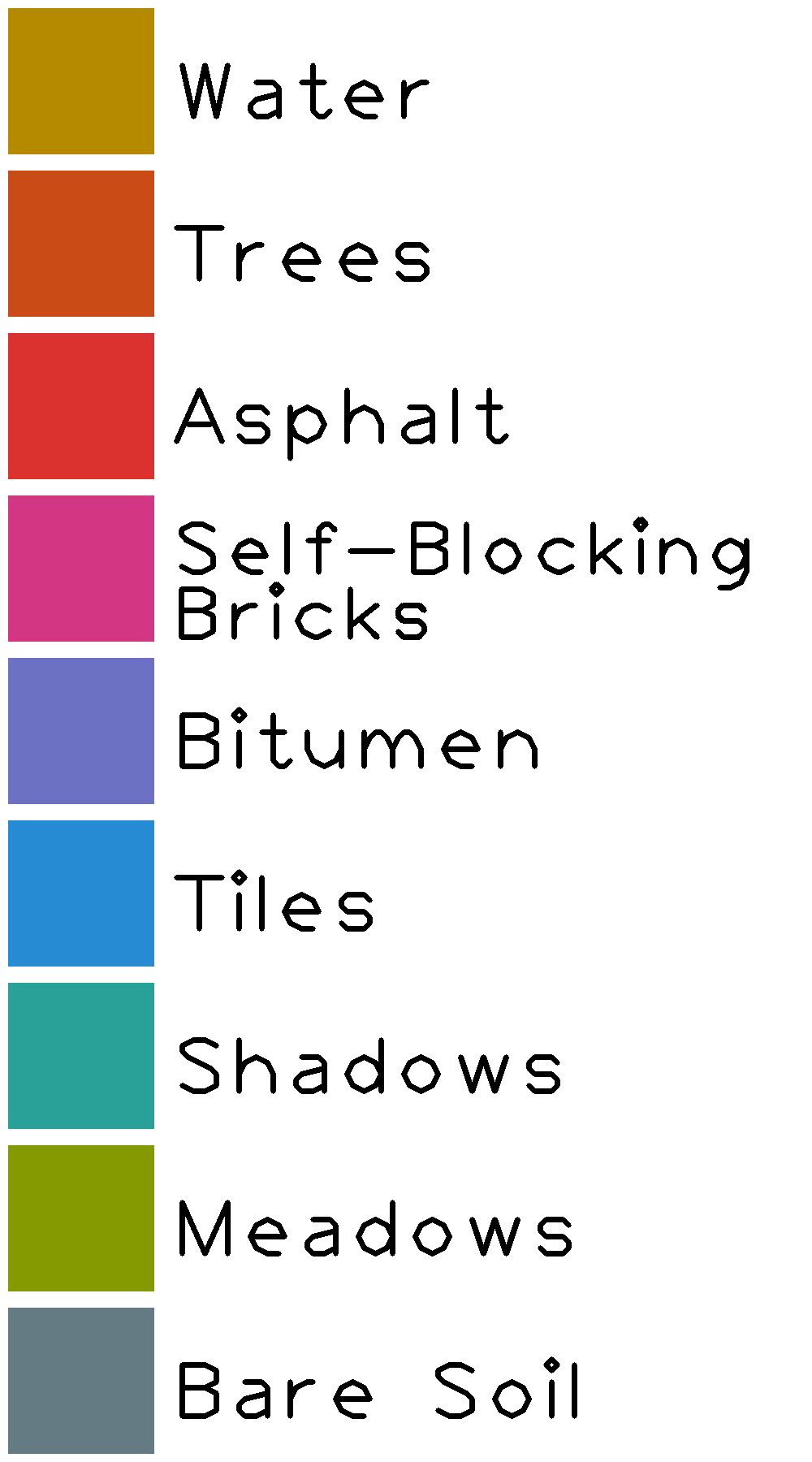}
	\caption{\label{paviac_fc_gt} The false color composite (band $10,20,40$) and groundtruth map of Pavia Centre} 
\end{figure}

The last image is the Pavia Centre scene, which is also returned by the ROSIS sensor over Pavia,
northern Italy. A sub-image of $1096\times 715$ pixels with $102$ relative
clean bands is taken into account, where the geometric resolution is $1.3$ meters. Ignoring the background pixels, the
groundtruth labels fall into $K=9$ reference classes, as given in \cref{label_paviac}.
The false color composite and the groundtruth map of Pavia Center are shown in \cref{paviac_fc_gt}.

\begin{table}[!hbp] \caption{\label{label_paviac}Reference classes and sizes of training and testing sets of Pavia Centre image}
	\centering
	\begin{tabular}{c|c|c|c|c}
		\hline
		\hline
		No. 	&Class	&Cardinality	&Train	&Test	\\
		\hline
		$1$	&Water	&$65971$	&$200$	&$65771$	\\
		$2$	&Trees	&$7598$	&$200$	&$7398$	\\
		$3$	&Asphalt	&$3090$	&$200$	&$2890$	\\
		$4$	&Self-Blocking	Bricks	&$2685$	&$200$	&$2485$	\\
		$5$	&Bitumen	&$6584$	&$200$	&$6384$	\\
		$6$	&Tiles	&$9248$	&$200$	&$9048$	\\
		$7$	&Shadows	&$7287$	&$200$	&$7087$	\\
		$8$	&Meadows	&$42826$	&$200$	&$42626$	\\
		$9$	&Bare	Soil	&$2863$	&$200$	&$2663$	\\
		\hline
		\multicolumn{2}{c|}{Total}&148152&1800&146352	\\
		\hline
		\hline
	\end{tabular}
\end{table}

\subsection{Experimental settings}\label{experiment}
\subsubsection{Training and testing sets}
Before further processing, each dataset is firstly normalized to have zero mean and unit
variance. To form the training set, $200$ pixels are randomly chosen from each class.
The two neuron networks, {\em i.e.}, the ANNC model and discriminant model, are trained by employing
the same training set. \cref{label_paviau,label_salinas,label_paviac} present
the sizes of training and testing sets in three datasets.

To train the ANNC model, the original training set is enlarged by virtual samples
 \cite{chen2016deep,guo2017spectral}, until the size of each class reaches $80000$. Using $x_1,x_2$
to denote two training pixels chosen from the same class $\ell$, a virtual sample $\tilde{x}$
with the same label is generated by
$\tilde{x} = q x_1+(1-q) x_2$, where $q$ is a random number chosen from the uniform distribution on $[-1,2]$.
The testing set is composed by all the unused pixels, which are directly fed to the learned ANNC model at
the testing stage.

To train the discriminant model in the spatial feature extraction algorithm, the pixel-pairs should be firstly generated. Let $Tr_i$ be the set of training pixels with
label $i$, for $i=1,2,\ldots,K$. The positive training set is expressed by
\begin{eqnarray*}
	Tr_+
	&=& \bigcup_{i} Tr_i \times Tr_i \\
	&=& \bigcup_{i} \{(x_1,x_2)|\,\forall x_1,x_2 \in Tr_i\}.
\end{eqnarray*}
The negative training set is generated by taking the Cartesian product between any two different classes of training pixels, namely
\begin{eqnarray*}
    Tr_-
    &=& \bigcup_{i\neq j} Tr_i\times Tr_j \\
    &=& \bigcup_{i\neq j} \{(x_1,x_2)|\,\forall x_1 \in Tr_i,\, x_2 \in Tr_j\}.
\end{eqnarray*}
Considering that the size of negative training set $Tr_-$ is overwhelmingly greater than
that of the positive set $Tr_+$, only half of the negative pixel-pairs are randomly chosen for training.
It is noteworthy that, the training pixels are excluded from either the testing samples or their neighbors on all the three datasets.

\subsubsection{Networks configurations}
The parameters in the ANNC network are set as recommended in~\cite{guo2017spectral}. To be precise, the weight of center loss is set to be $\lambda = 0.01$.
To train ANNC, the stochastic gradient descent (SGD) is applied with a mini-batch size $512$, and the learning rate is initialized by $0.01$ and decays by multiplying $0.3162$ every $20000$ steps.
This model is implemented on the
open source deep learning framework Caffe \cite{jia2014caffe}.

The discriminant model has a softmax layer at the top followed by the cross entropy loss function.
At the training stage, the SGD algorithm is used with the mini-batch size set to be $512$, where the learning rate is initialized by $0.01$ and decays every $50$ epochs by multiplying $0.1$.
For the convenience of using rectangular convolutional kernels,
this model is implemented on the compatible machine
learning framework Tensorflow \cite{abadi2016tensorflow}, following the codes of \cite{li2017hyperspectral}.
\subsubsection{Hyperparameter selection}
\begin{figure*}
	\centering
\graphicspath{{Figures/}}
	\subfigure[Pavia University]{
		\includegraphics[trim = 0mm 0mm 0mm 0mm, clip,width=0.37\textwidth]  {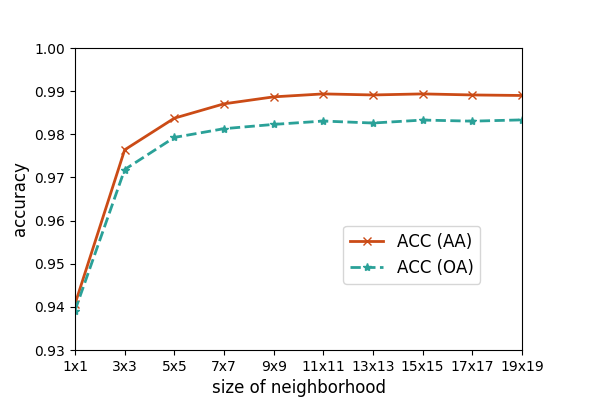}}
	\subfigure[Salinas]{
		\includegraphics[trim = 0mm 0mm 0mm 0mm, clip,width=0.37\textwidth]  {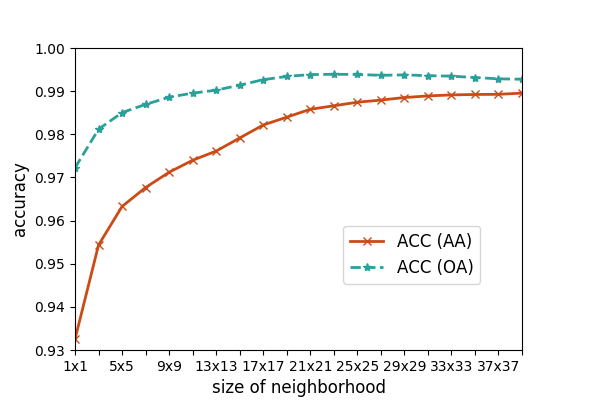}}
	\caption{\label{size_neib} Comparisons of classification accuracies in terms of AA and OA,  along with varying neighborhood sizes, on Pavia University and Salinas datasets. }
\end{figure*}

We discuss how to select the two crucial hyperparameters
introduced by
the proposed CSFF framework, {\em i.e.}, the size of neighborhood $N(x)$,
and the calibration threshold value $t$ in~\eqref{eq:binarymatrix}.
Firstly, to study the influence of neighborhood size on classification accuracies, we vary this parameter within some range, while fixing the threshold to a modest value with $t=0.01$. Experiments are performed on datasets Pavia University and Salines, with the neighborhood size varying within the sets \{$1 \times 1, 3 \times 3,...,19 \times 19$\} and \{$1 \times 1, 3 \times 3,...,39 \times 39$\}, respectively. 
As presented in~\figurename~\ref{size_neib}, an increasing neighborhood size generally leads to improvements on classification performance. This phenomenon is not surprising. Because of the good property of the convolutional kernel, only the useful spatial information from the neighboring spectra will contribute to the fusion of the spectral-spatial feature corresponding to the centering test pixel. A relatively larger neighborhood usually accounts for more useful spatial information, thus being favorable to boost the classification performance. However, it is also noticed that at the testing stage, a doubled neighborhood radius will quadruple the computational complexity.
By balancing the computational cost and the classification accuracy, the neighborhood size is set to be $19\times19$ for Pavia datasets and $39\times39$ for Salinas dataset.

\begin{figure*}
	\centering
\graphicspath{{Figures/}}
	\subfigure[Pavia University]{
		\includegraphics[trim = 0mm 0mm 0mm 0mm, clip,width=0.37\textwidth]  {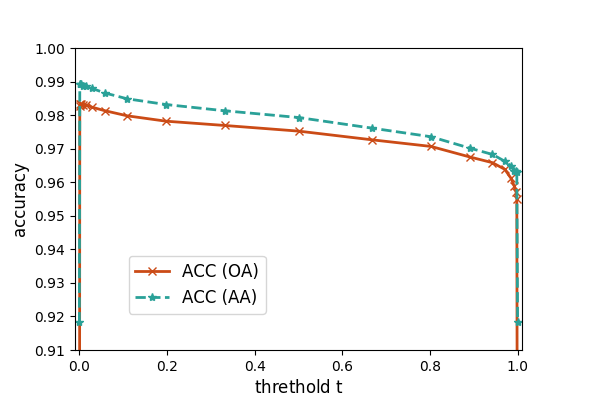}}
	\subfigure[Salinas]{
		\includegraphics[trim = 0mm 0mm 0mm 0mm, clip,width=0.37\textwidth]  {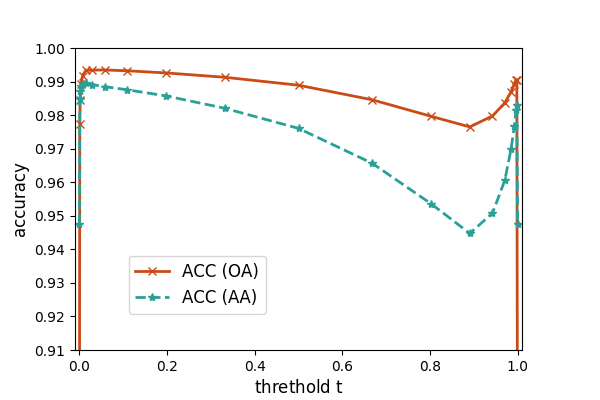}}
	\caption{\label{thsh} Comparisons of classification accuracies in terms of AA and OA, along with varying threshold values, on Pavia University and Salinas datasets.  }
\end{figure*}

Secondly, we validate the selection of threshold value $t$ in~\eqref{eq:binarymatrix}, which calibrates the probabilities in the spatial matrix into binary predictions. Experiments are performed on Pavia University and Salinas datasets by varying $t$ within interval $[0,1]$, where the neighborhood sizes are fixed as $19\times19$ and $39\times39$, respectively. As the learned discriminant model allows to estimate whether two paring pixels belong to the same class (probability near $1$) or not (probability near $0$), the elements presented in spatial matrices are generally distributed close to $0$ and $1$.
Thus, the classification performance tends to be particularly sensitive to the values of $t$ close to $0$ and $1$. Accounting for this fact, the threshold value $t$ is set by $t=\frac{1}{1+e^{-0.7x}}$,
where $x\in\mathbb{N}$ and $x\in[-9,11]$. Two extreme cases with $t=0$ and $t=1$ are also examined. It is noticed that with $t=0$, the proposed spectral-spatial feature fusion algorithm in~\eqref{fusion} is reduced to a simple average-over-neighborhood algorithm, while with $t=1$, the algorithm does not exploit any spatial information for feature fusion.
The classification results using different threshold values are given in \figurename{~\ref{thsh}}. As observed, on both datasets, small threshold values close to $t=0$ lead to promising classification accuracies, that are much more advantageous over the results obtained by $t=0$ and $t=1$. In this paper, we apply a unified threshold value with $t=0.01$ in all the experiments on three datasets.

\subsubsection{Classifiers}

To evaluate the performance of the spectral-spatial features $f_{N(x)}(x)$, a comparative study is performed by
classifying them using three classifiers, namely the center classifier mentioned in\cref{center_classifier},
the $k$NN algorithm \cite{altman1992introduction},
and the SVM algorithm \cite{chang2011libsvm}.
Without involving extra training data, these three classifiers are trained based on the same training pixels
as used in training the ANNC model and the discriminant model.
Precisely, by applying a transfer learning  strategy \cite{pan2010survey}, the classifiers are trained
with spectral features (with their respective labels) extracted by the learned ANNC model from
the training pixels. Here, the spectral-spatial features of the training pixels
are not directly used for training the classifiers,
for the sake that no additional information from the pixels other than the training ones should be used before
the testing stage. For a given training pixel $x$, its neighborhood $N(x)$ used for spectral-spatial feature generation
may contain the pixels from the testing set. Hence, the resulting spectral-spatial feature $f_{N(x)}(x)$ may contain the
information from the testing set, and is not proper for training the classifiers.
In practice,
the class centers in center classifier are estimated by averaging the spectral features of training pixels
within each class.
Concerning $k$NN, it assigns a test spectral-spatial feature to the class most common among its $k$ nearest
training spectral features, where two cases with $k=5$ and $k=10$ are considered in the experiments.
Similar to the $k$NN algorithm, the SVM algorithm is also trained by the spectral features with their respective labels.
The $\mathrm{rbf}$ kernel is applied and the multi-class classification is handled according to a one-vs-one scheme.
The Python package SciPy is applied directly for $k$NN and SVM algorithms, with the parameters,
that are not given explicitly here, set to
be the default values\footnote{The software is available at: https://www.scipy.org/.}.

\subsection{Results analysis}
We evaluate the classification performance of the proposed framework on the aforementioned datasets.
In this framework, the models are trained merely based on spectral pixels without using any
spatial information, as the latter is likely to coincide with the testing set. From this aspect,
three state-of-the-art methods are considered, which are the traditional method MASR \cite{Fang2014Spectral},
the CNN-based PPF \cite{li2017hyperspectral}, and the ANN-based ANNC-ASSCC \cite{guo2017spectral}.
As the state-of-the-art baseline of CNN models, the $3$D-CNN~\cite{chen2016deep} is also compared, which is trained based on pixel patches. For fair comparison,
all the comparative methods are trained using the training sets of same size,
namely $200$ random pixels for each class.

Three commonly-used metrics are adopted to evaluate the classification performance globally,
which are the overall accuracy (OA), the average accuracy (AA), and the Kappa coefficient ($\kappa$).
Briefly, OA represents the overall percentage of testing
samples that are correctly classified, while AA calculates the average value of the accuracies of
each class on testing samples.
The $\kappa$ coefficient measures the agreement between the predicted labels and groundtruth labels.

As illustrated in \cref{paviau,salinas,paviac},
competitive classification results are obtained by the proposed CSFF method on all the three datasets.
On the Pavia University scene, the CSFF provides the best classification accuracy in terms of OA, AA,
and $\kappa$, which signifies a $20\%-30\%$ drop in prediction failures over the second best ANNC-ASSCC.
Similar improvements are also observed on the Salinas scene, where the CSFF yields over $1.5\%$ increase
in OA when compared to the ANNC-ASSCC, the value corresponding to around $50\%$ fewer
prediction failures. Specifically, noteworthy improvements are achieved by CSFF
over the methods PPF and ANNC-ASSCC on addressing two difficult classification tasks, {\em i.e.,}
the Grapes untrained and the Vinyard untrained. It is noteworthy that although the MASR obtains
the best classification results, one possible reason is that the algorithm does not exclude the
training pixels when utilizing the spatial information, as explained in \cite{guo2017spectral}.
To this end, a modified version of MASR, termed MASR-t is applied on the Salinas scene for fairness,
as given in \cref{salinas}. Concerning the Pavia Centre scene, the CSFF still
leads to slight improvements on all the metrics, considering the high classification accuracies
achieved by the comparing counterparts.

\begin{table*}
	\centering
	\caption{\label{paviau}Classification accuracies (averaged over $5$ runs) of MASR, PPF,
		$3$D-CNN, ANNC-ASSCC and CSFF on Pavia University scene}
	\begin{tabular}{c|c|c|c|c|c}
		\hline \hline
		{}					 & MASR   		   & PPF    		 &3D-CNN	& ANNC-ASSCC & CSFF \\ \hline
		Asphalt              & $89.97\pm1.81$  & $97.25\pm0.35$  &$95.18\pm1.33$	& $98.69\pm0.78$      & $98.46\pm0.42$     \\ \cline{2-6}
		Meadows              & $98.78\pm0.36$  & $95.24\pm0.42$  &$98.90\pm0.28$	& $99.97\pm0.03$      & $99.79\pm0.03$     \\ \cline{2-6}
		Gravel               & $99.78\pm0.47$  & $94.17\pm0.49$  &$95.55\pm1.63$	& $93.85\pm2.15$      & $96.03\pm2.18$     \\ \cline{2-6}
		Trees                & $97.47\pm0.42$  & $97.20\pm0.30$  &$99.12\pm0.50$	& $96.68\pm0.99$      & $98.00\pm0.58$     \\ \cline{2-6}
		Painted metal sheets & $100.00\pm0.00$ & $100.00\pm0.00$ &$100.00\pm0.00$	& $100.00\pm0.00$     & $99.96\pm0.04$     \\ \cline{2-6}
		Bare Soil            & $99.87\pm0.23$  & $99.37\pm0.19$  &$98.23\pm1.14$	& $100.00\pm0.00$     & $100.00\pm0.00$    \\ \cline{2-6}
		Bitumen              & $100.00\pm0.00$ & $96.16\pm0.21$  &$91.04\pm5.60$	& $96.88\pm1.24$      & $98.91\pm0.33$     \\ \cline{2-6}
		Self-Blocking Bricks & $98.76\pm0.63$  & $93.83\pm0.59$  &$98.17\pm0.79$	& $93.11\pm2.50$      & $95.21\pm1.74$     \\ \cline{2-6}
		Shadows              & $92.17\pm1.78$  & $99.46\pm0.15$  &$99.68\pm0.56$	& $97.62\pm0.37$      & $100.00\pm0.00$    \\ \hline \hline
		OA($\%$)             & $97.42\pm0.36$  & $96.25\pm0.20$  &$98.14\pm0.10$	& \!\!\textcircled{\tiny2}$98.55\pm0.16$       & \!\!\textcircled{\tiny1}$98.90\pm0.14$     \\ \cline{2-6}
		AA($\%$)             & \!\!\textcircled{\tiny2}$97.42\pm0.40$  & $96.97\pm0.08$  &$97.32\pm0.68$	& \!\!\textcircled{\tiny2}$97.42\pm0.22$      & \!\!\textcircled{\tiny1}$98.49\pm0.13$     \\ \cline{2-6}
		$\kappa$             & $0.9654\pm0.0048$ & $0.9499\pm0.0026$ &$0.9687\pm0.0015$	& \!\!\!\textcircled{\tiny2}$0.9805\pm0.0021$     & \!\!\textcircled{\tiny1}$0.9852\pm0.0018$     \\ \hline \hline
	\end{tabular}
\end{table*}
\begin{table*}
	\centering
	\caption{\label{salinas}Classification accuracies (averaged over 5 runs)  of MASR, PPF,
		$3$D-CNN, ANNC-ASSCC and CSFF on Salinas}
	\begin{tabular}{c|c|c|c|c|c|c}
		\hline \hline
		& MASR   & MASR-t & PPF    & 3D-CNN & ANNC-ASSCC & CSFF \\ \hline
		Brocoli\_green\_weeds\_1     & \tabincell{c}{$100.00$\\$\pm0.00$} & \tabincell{c}{$100.00$\\$\pm0.00$} & \tabincell{c}{$99.98$\\$\pm0.03$} & \tabincell{c}{$100.00$\\$\pm0.00$}& \tabincell{c}{$100.00$\\$\pm0.00$}     & \tabincell{c}{$99.97$\\$\pm0.03$}     \\ \cline{2-7}
		Brocoli\_green\_weeds\_2     & \tabincell{c}{$99.97$\\$\pm0.05$}  & \tabincell{c}{$99.95$\\$\pm0.07$}  & \tabincell{c}{$99.58$\\$\pm0.09$} & \tabincell{c}{$99.64$\\$\pm0.39$} & \tabincell{c}{$100.00$\\$\pm0.00$}     & \tabincell{c}{$100.00$\\$\pm0.00$}    \\ \cline{2-7}
		Fallow                       & \tabincell{c}{$100.00$\\$\pm0.00$} & \tabincell{c}{$100.00$\\$\pm0.00$} & \tabincell{c}{$99.61$\\$\pm0.17$} & \tabincell{c}{$95.95$\\$\pm3.76$} & \tabincell{c}{$100.00$\\$\pm0.00$}     & \tabincell{c}{$100.00$\\$\pm0.00$}    \\ \cline{2-7}
		Fallow\_rough\_plow          & \tabincell{c}{$99.89$\\$\pm0.08$}  & \tabincell{c}{$99.66$\\$\pm0.18$}  & \tabincell{c}{$99.73$\\$\pm0.08$} & \tabincell{c}{$100.00$\\$\pm0.00$} & \tabincell{c}{$99.53$\\$\pm0.41$}      & \tabincell{c}{$95.54$\\$\pm3.64$}     \\ \cline{2-7}
		Fallow\_smooth               & \tabincell{c}{$99.58$\\$\pm0.14$}  & \tabincell{c}{$99.56$\\$\pm0.10$}  & \tabincell{c}{$97.43$\\$\pm0.28$} & \tabincell{c}{$99.92$\\$\pm0.10$} & \tabincell{c}{$99.67$\\$\pm0.17$}      & \tabincell{c}{$99.39$\\$\pm0.29$}     \\ \cline{2-7}
		Stubble                      & \tabincell{c}{$100.00$\\$\pm0.01$} & \tabincell{c}{$99.98$\\$\pm0.03$} & \tabincell{c}{$99.66$\\$\pm0.16$} & \tabincell{c}{$100.00$\\$\pm0.00$} & \tabincell{c}{$100.00$\\$\pm0.00$}     & \tabincell{c}{$99.93$\\$\pm0.05$}    \\ \cline{2-7}
		Celery                       & \tabincell{c}{$99.95$\\$\pm0.09$}  & \tabincell{c}{$99.97$\\$\pm0.02$}  & \tabincell{c}{$99.93$\\$\pm0.04$} & \tabincell{c}{$99.18$\\$\pm0.87$}& \tabincell{c}{$100.00$\\$\pm0.00$}     & \tabincell{c}{$99.94$\\$\pm0.05$}     \\ \cline{2-7}
		\textbf{Grapes\_untrained}            & \tabincell{c}{$97.82$\\$\pm0.46$} & \tabincell{c}{$87.72$\\$\pm0.84$}   & \tabincell{c}{$84.81$\\$\pm0.94$} & \tabincell{c}{$92.00$\\$\pm2.13$}& \tabincell{c}{$92.34$\\$\pm0.69$} & \tabincell{c}{$95.61$\\$\pm1.19$}     \\ \cline{2-7}
		Soil\_vinyard\_develop       & \tabincell{c}{$99.99$\\$\pm0.02$}  & \tabincell{c}{$99.99$\\$\pm0.01$} & \tabincell{c}{$99.15$\\$\pm0.78$} & \tabincell{c}{$98.72$\\$\pm0.69$} & \tabincell{c}{$99.99$\\$\pm0.01$}      & \tabincell{c}{$99.97$\\$\pm0.04$}     \\ \cline{2-7}
		Corn\_senesced\_green\_weeds & \tabincell{c}{$99.90$\\$\pm0.08$}  & \tabincell{c}{$99.59$\\$\pm0.36$}  & \tabincell{c}{$96.73$\\$\pm0.46$} & \tabincell{c}{$99.39$\\$\pm0.69$} & \tabincell{c}{$99.44$\\$\pm0.23$}      & \tabincell{c}{$98.98$\\$\pm0.15$}     \\ \cline{2-7}
		Lettuce\_romaine\_4wk        & \tabincell{c}{$100.00$\\$\pm0.00$} & \tabincell{c}{$100.00$\\$\pm0.00$} & \tabincell{c}{$99.45$\\$\pm0.15$} & \tabincell{c}{$100.00$\\$\pm0.00$} & \tabincell{c}{$100.00$\\$\pm0.00$}     & \tabincell{c}{$99.88$\\$\pm0.13$}     \\ \cline{2-7}
		Lettuce\_romaine\_5wk        & \tabincell{c}{$99.98$\\$\pm0.03$}  & \tabincell{c}{$99.96$\\$\pm0.05$} & \tabincell{c}{$100.00$\\$\pm0.00$}& \tabincell{c}{$100.00$\\$\pm0.00$} & \tabincell{c}{$100.00$\\$\pm0.00$}     & \tabincell{c}{$100.00$\\$\pm0.00$}    \\ \cline{2-7}
		Lettuce\_romaine\_6wk        & \tabincell{c}{$100.00$\\$\pm0.00$} & \tabincell{c}{$99.86$\\$\pm0.20$} & \tabincell{c}{$99.50$\\$\pm0.07$} & \tabincell{c}{$100.00$\\$\pm0.00$} & \tabincell{c}{$99.50$\\$\pm0.80$}      & \tabincell{c}{$99.09$\\$\pm1.08$}     \\ \cline{2-7}
		Lettuce\_romaine\_7wk        & \tabincell{c}{$99.90$\\$\pm0.24$}  & \tabincell{c}{$99.66$\\$\pm0.09$}  & \tabincell{c}{$99.47$\\$\pm0.30$} & \tabincell{c}{$100.00$\\$\pm0.00$}  & \tabincell{c}{$99.63$\\$\pm0.28$}      & \tabincell{c}{$98.00$\\$\pm1.28$}     \\ \cline{2-7}
		\textbf{Vinyard\_untrained}           & \tabincell{c}{$99.22$\\$\pm0.32$} & \tabincell{c}{$96.34$\\$\pm0.80$} & \tabincell{c}{$81.80$\\$\pm4.30$} & \tabincell{c}{$96.38$\\$\pm0.48$} & \tabincell{c}{$90.79$\\$\pm2.42$} & \tabincell{c}{$98.14$\\$\pm1.41$}     \\ \cline{2-7}
		Vinyard\_vertical\_trellis   & \tabincell{c}{$99.99$\\$\pm0.02$}  & \tabincell{c}{$99.59$\\$\pm0.50$}  & \tabincell{c}{$98.81$\\$\pm0.41$} & \tabincell{c}{$99.50$\\$\pm0.99$} & \tabincell{c}{$99.99$\\$\pm0.02$}      & \tabincell{c}{$99.95$\\$\pm0.10$}     \\ \hline \hline
		OA($\%$)                     & \tabincell{c}{\textcircled{\tiny1}$99.38$\\$\pm0.32$}  & \tabincell{c}{$96.74$\\$\pm0.80$}  & \tabincell{c}{$93.61$\\$\pm0.64$} & \tabincell{c}{$95.91$\\$\pm0.87$} & \tabincell{c}{$96.98$\\$\pm0.39$}      & \tabincell{c}{\textcircled{\tiny2}$98.53$\\$\pm0.34$}     \\ \cline{2-7}
		AA($\%$)                     & \tabincell{c}{\textcircled{\tiny1}$99.76$\\$\pm0.02$}  & \tabincell{c}{$98.86$\\$\pm0.50$}  & \tabincell{c}{$97.23$\\$\pm0.32$} & \tabincell{c}{$98.79$\\$\pm0.29$} & \tabincell{c}{$98.81$\\$\pm0.13$}      & \tabincell{c}{\textcircled{\tiny2}$99.02$\\$\pm0.20$}     \\ \cline{2-7}
		$\kappa$                     & \tabincell{c}{\textcircled{\tiny1}$0.9930$\\$\pm0.0009$} & \tabincell{c}{$0.9635$\\$\pm0.0007$} & \tabincell{c}{$0.9285$\\$\pm0.0072$} & \tabincell{c}{$0.9480$\\$\pm0.0108$} & \tabincell{c}{$0.9662$\\$\pm0.0044$}     &
		\tabincell{c}{\textcircled{\tiny2}$0.9835$\\$\pm0.0038$}     \\ \hline \hline
	\end{tabular}
\end{table*}
\begin{table*}[!htb]
	\centering
	\caption{\label{paviac}Classification accuracies (averaged over 5 runs)  of MASR, PPF,
		$3$D-CNN, ANNC-ASSCC and CSFF on Pavia Centre}
	\begin{tabular}{c|c|c|c|c|c}
		\hline \hline
		& MASR   		   & PPF    		 &  3D-CNN			& ANNC-ASSCC & CSFF \\ \hline
		Water                & $99.87\pm0.11$  & $99.15\pm0.18$  &$99.93\pm0.11$	& $100.00\pm0.00$     & $100.00\pm0.01$    \\ \cline{2-6}
		Trees                & $94.22\pm0.45$  & $97.96\pm0.24$  &$98.26\pm0.31$	& $98.75\pm0.62$      & $98.75\pm0.66$     \\ \cline{2-6}
		Asphalt              & $99.45\pm0.46$  & $97.37\pm0.18$  &$94.98\pm1.97$	& $99.26\pm0.16$      & $98.61\pm0.48$     \\ \cline{2-6}
		Self-Blocking Bricks & $99.98\pm0.05$  & $99.27\pm0.11$  &$98.48\pm2.11$	& $99.96\pm0.04$      & $99.93\pm0.02$     \\ \cline{2-6}
		Bitumen              & $98.75\pm0.51$  & $98.79\pm0.16$  &$99.67\pm0.18$	& $99.35\pm0.28$      & $99.69\pm0.20$     \\ \cline{2-6}
		Tiles                & $80.23\pm1.78$  & $98.95\pm0.08$  &$99.29\pm0.41$	& $99.73\pm0.06$      & $99.56\pm0.09$     \\ \cline{2-6}
		Shadows              & $99.34\pm0.47$  & $94.36\pm0.35$  &$97.88\pm0.97$	& $97.49\pm0.64$      & $98.21\pm0.69$     \\ \cline{2-6}
		Meadows              & $99.90\pm0.04$  & $99.90\pm0.03$  &$99.99\pm0.01$	& $99.41\pm0.06$      & $99.90\pm0.02$     \\ \cline{2-6}
		Bare Soil            & $84.80\pm1.02$  & $99.96\pm0.05$  &$99.15\pm0.75$	& $98.72\pm0.56$      & $99.99\pm0.02$     \\ \hline \hline
		OA($\%$)             & $98.02\pm0.12$  & $99.03\pm0.08$  &$99.60\pm0.07$	& \textcircled{\tiny2}$99.73\pm0.05$      & \textcircled{\tiny1}$99.75\pm0.07$     \\ \cline{2-6}
		AA($\%$)             & $95.17\pm0.20$  & $98.41\pm0.06$  &$98.58\pm0.47$	& \textcircled{\tiny2}$99.25\pm0.09$      & \textcircled{\tiny1}$99.40\pm0.13$     \\ \cline{2-6}
		$\kappa$             & $0.9719\pm0.0018$ & $0.9862\pm0.0011$ &$0.9845\pm0.0010$ & \textcircled{\tiny2}$0.9937\pm0.0011$     & \textcircled{\tiny1}$0.9964\pm0.0009$
		\\ \hline \hline
	\end{tabular}
\end{table*}

The proposed CSFF generates explicit spectral-spatial features that
can be classified using different classification algorithms.
To examine the effectiveness of the features obtained by the CSFF under
various classifiers, experiments are performed by using the default center classifier,
$k$NN, and SVM, on all
the datasets. The results are given in \cref{paviau_c,salinas_c,paviac_c}. We observe that on all the images, the three classifiers yield stable and
similar classification accuracies in terms of all the quantitative metrics. It demonstrates that
the spectral-spatial features generated by the proposed framework are effective and robust
against different classifiers.

\begin{table*}
	\centering
	\caption{\label{paviau_c}Classification accuracies (averaged over $5$ runs) of $k$NN, SVM and center classifier on
	 spectral-spatial feature of Pavia University scene}
	\begin{tabular}{c|c|c|c|c}
		\hline \hline
		& $k$NN ($5$) & $k$NN ($10$) & SVM    & C-Classifier \\ \hline
		OA($\%$)             & $98.92\pm0.19$      & $98.93\pm0.23$       & $98.91\pm0.12$  & $98.90\pm0.14$        \\ \cline{2-5}
		AA($\%$)             & $98.54\pm0.23$     & $98.53\pm0.25$       & $98.54\pm0.10$  & $98.49\pm0.13$        \\ \cline{2-5}
		$\kappa$             & $0.9857\pm0.0026 $    & $0.9858\pm0.0031$      & $0.9856\pm0.0015$ & $0.9852\pm0.0018$       \\ \hline \hline
	\end{tabular}
\end{table*}
\begin{table*}
	\centering
	\caption{\label{salinas_c}Classification accuracies (averaged over 5 runs)  of $k$NN, SVM and center classifier on spectral-spatial feature of Salinas scene}
	\begin{tabular}{c|c|c|c|c}
		\hline \hline
		& $k$NN ($5$) & $k$NN ($10$) & SVM    & C-Classifier \\ \hline
		OA($\%$)                     & $96.80\pm0.94$      & $97.56\pm0.77$       & $98.42\pm0.39$  & $98.53\pm0.34$        \\ \cline{2-5}
		AA($\%$)                     & $98.36\pm0.52$      & $98.68\pm0.32$       & $98.92\pm0.24$  & $99.02\pm0.20$         \\ \cline{2-5}
		$\kappa$                     & $0.9642\pm0.0105$     & $0.9727\pm0.0087$      & $0.9823\pm0.0044$ & $0.9835\pm0.0038$        \\ \hline \hline
	\end{tabular}
\end{table*}
\begin{table*}
	\centering
	\caption{\label{paviac_c}Classification accuracies (averaged over 5 runs)  of $k$NN, SVM and center classifier on spectral-spatial feature of Pavia Centre scene}
	\begin{tabular}{c|c|c|c|c}
		\hline \hline
		& $k$NN ($5$) & $k$NN ($10$) & SVM    & C-Classifier \\ \hline
		OA($\%$)             & $99.77\pm0.07$       & $99.75\pm0.08$        & $99.74\pm0.07$  & $99.75\pm0.07$        \\ \cline{2-5}
		AA($\%$)             & $99.45\pm0.13$       & $99.41\pm0.16$        & $99.40\pm0.14$  & $99.40\pm0.13$        \\ \cline{2-5}
		$\kappa$             & $0.9967\pm0.0010$      & $0.9965\pm0.0011$       & $0.9963\pm0.0010$ & $0.9964\pm0.0009$       \\ \hline \hline
	\end{tabular}
\end{table*}

\subsection{Computational Cost}
The testing time of all the comparing methods are reported in~\tablename~\ref{Tab.Time}. Experiments are performed on a machine equipped with CPU of Intel Xeon E5-2660@2.6GHz and GPU of NVIDIA TitanX. As for the proposed CSFF, both the total testing time and the time of applying discriminant model are presented.
In fact, it is the testing data preparation in discriminant model, namely the generation of pixel-pairs using the centering pixel and each spectrum within the neighborhood, that is the most time-consuming. To alleviate computational burden, a slightly decreased neighborhood size can be adopted in practice. On one hand, this does not deteriorate the classification accuracies too much, as shown in \figurename{~\ref{size_neib}}. On the other hand, a declined neighborhood size reduces the computational complexity of testing data preparation, which computes $O(n^2)$ with $n$ being the neighborhood size.

\begin{table*}[!ht]
	\centering
	\caption{Comparison of testing time (in seconds)}\label{Tab.Time}
	\begin{tabular}{c|c|c|c|c|c}
		\hline \hline
		{}      & MASR   & PPF   &  3D-CNN  &  ANNC-ASSCC & CSFF (\textsl{Discriminant Model})\\
		\hline
		Pavia University &$2424.09$&$13.85$ & $50.94$ &$37.70$ & $1779.45$ ($\textsl{1771.02}$)\\
		\hline
		Salinas &$10572.88$&$22.51$&$45.69$ & $61.09$ & $5453.36$ ($\textsl{5436.76}$)\\
		\hline
		Pavia Centre &$8480.59$& $92.66$  & $164.06$ &  $135.87$ & $5924.74$ ($\textsl{5897.63}$)\\
		\hline \hline
	\end{tabular}
\end{table*}

\section{Conclusion}\label{sec: Conclusion}

In this paper, we investigated a novel ANN and CNN based classification framework that properly integrates the
spatial information to the spectral-based features, and generates spectral-spatial features suitable for
various classifiers. Based on a limited number of labeled pixels and without using any local information, both
a spectral feature extraction model and a discriminant model were trained. Using the learned discriminant model,
the local structure was extracted and represented as a customized convolutional kernel.
The spectral-spatial feature was obtained by a convolutional operation between the kernel and the corresponding
spectral features within a neighborhood. Experiments on three real hyperspectral images validated the performance
of the proposed method in terms of classification accuracies, when compared to the state-of-the-art algorithms.
We also studied the characteristics of the learning features, which showed robustness and stableness against
various classifiers. Future works will focus on extending the proposed framework to multiple-model fusion.


\bibliographystyle{IEEEtran}
\bibliography{AlanGuo,bib_fei}

\end{document}